\documentclass[12pt]{article}

\usepackage[top=2cm, bottom=2cm, left=1.5cm, right=1.5cm]{geometry}
\usepackage{natbib}

\usepackage{graphicx}%
\usepackage{multirow}%
\usepackage{amsmath,amssymb,amsfonts}%
\usepackage{amsthm}%
\usepackage{mathrsfs}%
\usepackage[title]{appendix}%
\usepackage{xcolor}%
\usepackage{textcomp}%
\usepackage{manyfoot}%
\usepackage{booktabs}%
\usepackage{listings}%

\usepackage{latexsym}

\usepackage{url}
\usepackage{xcolor}

\usepackage{rotating}

\usepackage{latexsym}
\usepackage{mathabx}

\usepackage{url}
\usepackage{lineno}
\usepackage{arydshln}

\usepackage{bm}

\usepackage{xspace}
\newcommand{\ie}{\emph{i.e.}\xspace}

\newcommand{\eg}{\emph{e.g.}\xspace}

\newcommand{\espace}{\vspace{7pt}}

\newcommand{\T}[1]{\ensuremath{\mathcal{#1}}} 
\newcommand{\M}[1]{\ensuremath{\bm{#1}}} 
\newcommand{\V}[1]{\ensuremath{\bm{#1}}} 

\newcommand{\swotted}{{\sc SWoTTeD}\xspace} 

\usepackage{algorithm2e}
\usepackage{lscape}
\usepackage{hyperref}

\SetKwInput{KwNotation}{Definition}

\graphicspath{{./Figures/}}

\title{\swotted: An Extension of Tensor Decomposition \\to Temporal Phenotyping}
\author{Hana Sebia$^\dagger$, Thomas Guyet$^\dagger$ and Etienne Audureau$^\ddagger$}
\date{
$\dagger$. AIstroSight, Inria, Hospices Civils de Lyon, Université Claude Bernard Lyon 1, \\
56 Bld Niels Bohr, 69603 Villeurbanne, France\\
\href{mailto:hana.sebia@inria.fr}{hana.sebia@inria.fr}\\
$\ddagger$. CEPIA Team, Paris University Hospital, Créteil, France.
}

\begin{document}

\maketitle
\abstract{Tensor decomposition has recently been gaining attention in the machine learning community for the analysis of individual traces, such as Electronic Health Records (EHR). 
However, this task becomes significantly more difficult when the data follows complex temporal patterns. 
This paper introduces the notion of a temporal phenotype as an arrangement of features over time and it proposes \swotted (\textbf{S}liding \textbf{W}ind\textbf{o}w for \textbf{T}emporal \textbf{Te}nsor \textbf{D}ecomposition), a novel method to discover hidden temporal patterns. \swotted integrates several constraints and regularizations to enhance the interpretability of the extracted phenotypes. 
We validate our proposal using both synthetic and real-world datasets, and we present an original usecase using data from the Greater Paris University Hospital. The results show that \swotted achieves at least as accurate reconstruction as recent state-of-the-art tensor decomposition models, and extracts temporal phenotypes that are meaningful for clinicians.}

\section{Introduction}
A tensor is a natural representation for multidimensional data. 
Tensor decomposition is a historic statistical tool for analyzing such complex data. 
The popularization of efficient and scalable machine learning techniques has made them attractive for real-world data~\citep{perros2017spartan}. 
It has therefore been successfully investigated in a number of areas, such as signal processing, chemometrics, neuroscience, communication or psychometrics \citep{FANAEET2016130}. 
Technically, tensor decomposition simplifies a multidimensional tensor into simpler tensors by learning latent variables in an unsupervised fashion~\citep{anandkumar2014tensor}. 
Latent variables are unobserved features that capture hidden behaviors of a system. 
Such variables are difficult to extract from complex multidimensional data due to 1) multiple interactions between dimensions and 2) intertwined occurrences of hidden behaviors. 

Recently, several approaches based on tensor decomposition have shown their effectiveness and their interest for computational phenotyping from Electronic Health Records (EHR)~\citep{Becker}. 
The hidden recurrent patterns that are discovered in these data are called \textit{phenotypes}. 
These phenotypes are of particular interest to 1) describe the real practices of medical units and 2) support hospital administrators to improve their care management. 
For example, a better description of care pathways of COVID-19 patients at the beginning of the pandemic may help clinicians to improve care management of future epidemic waves. This example motivates to apply such data analytic tools on a cohort of patients from the Greater Paris University Hospitals (see the case study in Section~\ref{sec:casestudy}).

The main limitation of existing tensor decomposition techniques is the definition of a phenotype as a mixture of medical events without considering the temporal dimension. 
This means that all events occur at the same time. 
In this case, a care pathway is viewed as a succession of independent daily cares. 
Nonetheless, it seems more realistic to interpret a care pathway as mixtures of \textit{treatments}, \ie sequences of cares. 
For example, COVID-19 patients hospitalized with acute respiratory distress syndrome are treated for several problems during the same visit: viral infection, respiratory syndromes and hemodynamic problems. 
On the one hand, a treatment of the viral infection involves the administration of drugs for several days. On the other hand, the acute respiratory syndrome also requires continuous monitoring for several days. A patient's care pathway can then be abstracted as a mixture of these treatments. 
Some approaches proposed to capture the temporal dependencies between daily phenotypes using temporal regularization~\citep{CNTF_2019} but the knowledge provided to the clinician are still daily phenotypes.

In this article, we present \swotted (\textbf{S}liding \textbf{W}ind\textbf{o}w for \textbf{T}emporal \textbf{Te}nsor \textbf{D}ecomposi\-tion), a tensor decomposition technique based on machine learning to extract temporal phenotypes.  
Contrary to a classical daily phenotype, a temporal phenotype describes the arrangement of drugs/procedures over a time window of several days. 
Drawing a parallel with sequential pattern mining, the state-of-the-art methods extract itemsets from sequences while \swotted extracts sub-sequences.
Thus, temporal phenotyping significantly enhances the expressivity of computational phenotyping. 
Following the principle of tensor decomposition, \swotted discovers temporal phenotypes that accurately reconstruct an input tensor with a time dimension. 
It allows the overlapping of distinct occurrences of phenotypes to represent asynchronous starts of treatments. 
To the best of our knowledge, \swotted is the first extension of tensor decomposition to temporal phenotyping. 
We evaluate the proposed model using both synthetic and real-world data. 
The results show that \swotted outperforms the state-of-the-art tensor decomposition models in terms of reconstruction accuracy and noise robustness. 
Furthermore, the qualitative analysis shows that the discovered phenotypes are clinically meaningful. 

\espace

In summary, our main contributions are as follows:
\begin{enumerate}
\item We extend the definition of tensor decomposition to temporal tensor decomposition. To the best of our knowledge, this is the first extension of tensor factorization that is capable of extracting temporal  patterns. A comprehensive review is provided to position our proposal within the existing approaches in different fields of machine learning.
\item We propose a new framework, denoted as \swotted, for extracting temporal phenotypes through the resolution of an optimization problem. This model also introduces a novel regularization term that enhances the quality of the extracted phenotypes. \swotted has been extensively tested on synthetic and real-world datasets to provide insights into its competitive advantages. Additionally, we offer an open-source, well-documented, and efficient implementation of our model.
\item We demonstrate the utility of temporal phenotypes through a real-world case study. 
\end{enumerate}

\espace

The remainder of the article is organized as follows: the next section presents the state of the art of machine learning techniques related to tensor decomposition in the specific case of temporal tensor. Section~\ref{sec:problem} introduces the new problem of temporal phenotyping, then Section~\ref{sec:swotted} presents \swotted to solve it. The evaluation of this model is detailed in three sections. We begin by introducing the experimental setup in Section~\ref{sec:setup}, followed by the presentation of reproducible experiments and results conducted on synthetic and real-world datasets. Lastly, Section~\ref{sec:casestudy} presents a case study on a COVID-19 dataset.


\section{Related Work}\label{sec:soa}

Discovering hidden patterns, a.k.a \textit{phenotypes}\footnote{The term of phenotype usually denotes a set of traits that characterizes a disease. This notion is here extended to traits observed through the EHR systems. Thus, a phenotype is a set of observations in EHR data characterizing a treatment or a disease.} in our work, from longitudinal data is a fundamental issue of data analysis. 
This problem has been more especially investigated for the analysis of EHR data which are complex and require to be explored to discover hidden patterns providing insights about patient cares. 
With this objective, tensor decomposition has been widely used and proven to extract concise and interpretable patterns~\citep{Becker}. 

In this related work, we enlarge the scope and also review different machine learning techniques that have been recently proposed to address the problem of patient phenotyping. 
As our proposal is based on the principles of tensor decomposition, we start by reviewing techniques derived from tensor decomposition and that have been applied to EHR. 
Then, we present methods that targeted a similar objective, but with alternative modeling techniques. 
They both share the task of uncovering hidden patterns in temporal sequences using unsupervised methods. 

\paragraph{Notations} In the remaining of this article, $[K]=\{1,\dots,K\}$ denotes the set of the $K$ first non-zero integers. $\mathbb{N}^*$ denotes the strictly positive natural numbers.
Curvy capital letters ($\T{X}$) denote tensors (or irregular tensors), bold capital letters ($\M{X}$) are matrices, bold lowercase letters ($\V{x}$) are vectors and lowercase letters ($x$) are scalars. 

\subsection{Tensor Decomposition from Temporal EHR Data}

An EHR dataset has a timed event-based structure described at least by  three dimensions: patient identifiers, care events (procedures, lab tests, drug administrations, etc) and time. 
We begin by introducing the tensor-based representation of EHR data. 
Following that, we provide a comprehensive review of various techniques designed to tackle the challenge of patient phenotyping using this data representation.

\subsubsection{Tensor Based Representation of Temporal EHR Data}\label{sec:soa:tensorbasedrepresentation}

Considering that time is discrete (\eg, events are associated with a specific day during the patient's stay), each patient $k\in[K]$ is represented by a matrix $\M{X}^{(k)}$ where the first dimension represents the  type of events and the second one represents time.
If patient~$k$ received a care event~$i$ at time~$t$, then $x^{(k)}_{i,t}$ is a non-zero value. 
In the majority of cases, values are categorical, typically represented as boolean values (0 or 1). However, there are cases where  values may be integers or real numbers, such as the count of drugs administered or a measurement of a biophysiological parameter.

\begin{figure}[tb]
\centering
\includegraphics[width=.75\textwidth]{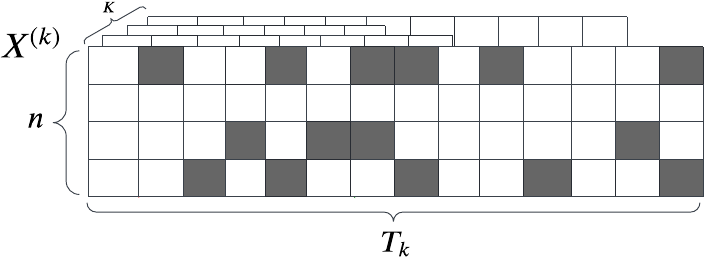}
\caption{Illustration of an irregular tensor $\T{X}=\{\M{X}^{(k)}\}_{k\in[K]}$ representing a collection of $K$ patients stays. Each patient has its own duration $T_k$ but share the same set of cares (in rows). A black cell at position $(i,t)$ (\ie $x^{(k)}_{i,t}= 1$) indicates that the $i$-th care occurs at the time~$t$.}
\label{fig:tensor}
\end{figure}

Additionally, if we consider that all patients have the same length of stay, the set of patients is a regular three-dimensional tensor~$\T{X}$, \ie a data-cube. 
Nevertheless, in practice, patients' stays do not have the same duration. 
It ensues that each matrix~$\M{X}^{(k)}$ has its own temporal size, noted $T_k$. In this case, the collection of matrices~$\{\M{X}^{(k)}\}_{k\in[K]}$ can not be stacked as a regular third-order tensor. 
Such a collection is termed an \textit{irregular tensor} and we use the same notation~$\T{X}$.
Figure~\ref{fig:tensor} depicts an irregular tensor that represents the typical structure of the input of a patient phenotyping problem. 
In this figure, we assume all features are categorical, \ie matrices are boolean valued. A black cell represents a 1 (the presence of a given event at a given time instant) and a white cell represents a 0 (the absence of a given event at a given time instant).

We will see that some tensor decomposition approaches handle irregular tensors while some others require regular ones. 
Padding the shortest care pathways with zeros to conform to a regular standardized tensor structure is an appealing option. However, two primary drawbacks come with this approach: 1) It would artificially inflate dataset sizes when accommodating a single lengthy care pathway; 2) It would blur the distinction between the absence of an event during a hospital stay and non-hospital stays, potentially undermining result accuracy and interpretability. For these reasons, irregular tensors are more suitable to represent care pathways.

\subsubsection[]{Third-order Tensor Decomposition with PARAFAC2}
PARAFAC2 \citep{kiers1999parafac2} is a seminal decomposition designed to handle irregular tensors. 
It extends the Canonical Polyadic (CP) factorization \citep{kiers1999parafac2} which is a foundational approach for decomposing regular tensors into a sum of rank-one components. 
PARAFAC2 introduces flexibility in accommodating tensors with different temporal lengths, making it particularly useful in scenarios where temporal information is crucial, such as in the analysis of longitudinal data~\citep{fanaee2015eigenevent} and specifically electronic health records~\citep{PARAFAC_2019}.

Recent enhancements to PARAFAC2 have significantly strengthened the capabilities of this model to address specific challenges. 
In particular, SPARTan, proposed by~\cite{perros2017spartan}, stands out for its scalability and parallelizability on large and sparse datasets, while Dpar2~\citep{jang2022dpar2} was designed to handle effectively irregular dense tensors. 
Nonetheless, these interesting computational properties do not ensure that the solution given by PARAFAC2 identifies meaningful phenotypes in practice. 
For instance, factor matrices can contain negative values and this does not make sense in the context of patient phenotyping. 

Alternative formulations of PARAFAC2 have been proposed to incorporate additional constraints. \cite{cohen2018nonnegative} introduced a non-negativity constraint on the varying mode to enhance the interpretability of the resulting factors. \cite{roald2022admm} extended the constraints to all-modes. 
On the other hand, COPA (Constrained PARAFAC2)~\citep{COPA2018} took a step further by introducing various meaningful constraints in PARAFAC2 modeling, including latent components that change smoothly over time and sparse phenotypes to ease the interpretation.

Finally, a practical limitation of PARAFAC2 pertains to the rank value. In the context of patient phenotyping, the rank value represents the number of phenotypes. The decomposition technique requires that this value must not exceed the dimension of any mode, including the time dimension. 
Consequently, the number of phenotypes cannot exceed the minimum duration observed in an irregular third-order tensor, which implies the exclusion of pathways with duration shorter than the specified rank from the dataset.

In conclusion, PARAFAC2 proves to be an interesting model for patient phenotyping using temporal EHR data. It is especially suitable to handle large datasets of irregular tensors such as care pathways. However, its formulation may lead to extract meaningless phenotypes. To address this limitation, some extensions of PARAFAC2 include additional constraints, but they remain simple.  
The extraction of more meaningful and clinically relevant phenotypes requires flexible constraints.

\subsubsection{Extracting Relevant Phenotypes with Tensor Decomposition}
Recently, several approaches have been proposed to enrich tensor decomposition with expert constraints designed to yield more meaningful and clinically relevant phenotypes. 
We identified three specific questions that have been addressed in the literature:
\begin{description}
\item[Dealing with additional static information] 
Real EHR data contain both temporal (\eg, longitudinal clinical visits) and static information (\eg, age, body mass index (BMI), smoking status, main reason for hospitalisation, etc.). 
It is expected that the static information impacts the temporal phenotypes, \ie the distribution of care deliveries along the patient visits.
This question has been addressed by TASTE~\citep{Afshar2020TASTETA} and TedPar~\citep{yin2021tedpar}. 
TASTE takes as input an irregular tensor $\T{X}$ and an additional matrix representing the static features of each patient. The decomposition maps input data into a set of phenotypes and patients' temporal evolution. Phenotypes are defined by two factor matrices: one for temporal features and and the other for static features.

\item[Dealing with correlations between diagnoses and medication events] 
In general, the set of medical events contains diagnostic events (e.g., lab tests) and care events (e.g., medications or medical procedures). The occurrences of these two types of events may be correlated: it is likely that establishing a diagnosis leads to the delivery of a specific care. 
For instance, an elevated blood glucose concentration, measured by a lab test, often leads to the administration of insulin, which constitutes a care event.
In patient data, this implies that the likelihood of a blood glucose measurement and an insulin injection co-occurring is higher than, for example, a blood glucose measurement and mechanical ventilation.
HITF~\citep{yin2018joint} leverages these correlations to enhance tensor decomposition by splitting the dimension of feature types into two dimensions: one for medications and the other for diagnoses.
This model has also been used within CNTF~\citep{CNTF_2019}, which proposes modeling patient data as a tensor with four dimensions: patients identifier, lab tests, medications, and time.

\item[Supervision of tensor decomposition] Another improvement of the tensor decomposition methods involves extending them to supervised fashions. 
\cite{henderson2018phenotyping} proposed a semi-supervised tensor factorization method that introduces a cannot-link matrix on the patient factor matrix to encourage separation in the patient subgroups.
 
Predictive Task Guided Tensor Decomposition (TaGiTeD) \citep{yang2017tagited} is another framework conceived to overcome the limitations of existing unsupervised approaches, such as the requirement for a large dataset to achieve meaningful results. TaGiTeD guide the decomposition by specific prediction tasks. This is done by learning representations that lead to best prediction performances. 
Lastly, Rubik \citep{wang2015rubik} and SNTF \citep{anderson2017supervised} are other tensor factorization models incorporating guidance constraints to align with existing medical knowledge.
\end{description}

It is worth noting that these advanced models benefit from the recent progress in machine learning, and more specifically automatic differentiation~\citep{baydin2018automatic}. 
Automatic differentiation does not necessitate explicit gradient computation to evaluate efficiently the derivatives of a function. 
Thus, it eases the design of efficient optimization algorithms for various tensor decomposition tasks. 
The flexibility of these machine learning frameworks fosters the conception of complex models that produce more meaningful phenotypes.

\subsubsection{Temporal Dimension in Tensor Decomposition}\label{sec:soa:temp}

While it appears crucial to manage the dynamics in a patient's evolution, most tensor decompositions do not explicitly model the temporal dependencies within the patient data. The temporal aspect is particularly significant when constructing phenotypes for typical care profiles.

First of all, it is worst noting that the seminal decomposition model, PARAFAC2, does not capture temporal information in their phenotypes. 
Let $\T{X}=\left(\M{X}^{(k)}\right)_{k\in[K]}$ be an irregular tensor of $K$ patients, and $\T{Y}=\left(\M{Y}^{(k)}\right)_{k\in[K]}$ another irregular tensor such that $y_{:,t}^{(k)}=x_{:,\rho_k(t)}^{(k)}$ where $\rho_k$ is a random permutation of daily vectors of the patient $k$. 
The decomposition of these two tensors leads to the exact same phenotypes. 
This illustrates that the phenotypes extracted by PARAFAC2 are insensitive to the temporal dimension of the data.

Some variants of PARAFAC2 have targeted this limitation by the introduction of temporal regularisation terms. According to~\cite{COPA2018}, learning temporal factors that change smoothly over time is often desirable to improve the interpretability and alleviate the over-fitting.
COPA, as introduced by ~\cite{COPA2018}, includes a smoothness regularization to account for irregularities in the temporal gaps between two visits. LogPar~\cite{yin2020logpar} expands this regularization to binary and incomplete irregular tensors. However, both of these techniques have limitations as they rely solely on local information to smooth pathways, neglecting the temporal history needed to construct meaningful phenotypes.

To address long term dependencies, Temporally Dependent PARAFAC2 Factorization (TedPar) \citep{yin2021tedpar} was developed to model the gradual progression of chronic diseases over an extended period. TedPar introduces the concept of temporal transitions from one phenotype to another to capture temporal dependencies. Additionally, \cite{AhnJK22} proposed Time-Aware Tensor Decomposition (TATD), a tensor decomposition method that incorporates time dependency through a smoothing regularization with a Gaussian kernel.
For CNTF~\citep{CNTF_2019}, a recurrent neural network (RNN) was used to take into account the ordering of the clinical events. Given the sequence $w^{k}_{p,1},\ldots, w^{k}_{p,t-1} $ describing the progression of a phenotype $p$ of a given patient $k$, an LSTM (Long Short-Term Memory) network is used to predict $w^{k}_{p,t}$ such that the Mean Square Error (MSE) between the real and predicted value is minimized. 
The idea behind this is to penalize a reconstruction model that does not allow to accurately predict the next sequence of events. It enforces to discover a decomposition that is easily predictable.

\espace

It is worth noting that all these models do not discover temporal patterns. The temporal dimension is used to constraint the extraction of daily phenotypes by taking into account the temporal dependencies.
Nonetheless, these temporal dependencies are not explicit for a physician analyzing the care pathways. 
A daily phenotype shown to physicians only represent co-occurring events. 
The method presented in this article extracts phenotypes that describe a temporal pattern. The phenotype itself encapsulates information about temporal dependencies in an easily interpretable manner.

\subsection{Alternative Approaches for Extracting Temporal Phenotypes}

While tensor decomposition techniques have not yet tackled the issue of extracting temporal patterns from care pathways, similar challenges have been addressed using alternative approaches. In this section, we highlight three of them: 

\begin{description}
\item[Temporal Extensions of Topic Models]
Originally, topic modeling (or latent block models) is a statistical technique for discovering the latent semantic structures in textual document. It can estimate, at the same time, the mixture of words that is associated with each topic, and the mixture of topics that describes each document.  
\cite{PIVOVAROV2015156} and \cite{ahuja2022mixehr} proposed to consider the patients' data as a collection of documents. The topic modeling of these documents results in a set of topics representing the phenotypes. 
Temporal extensions of topic modeling could then be used to extract temporal phenotypes. For instance, Temporal Analysis of Motif Mixtures (TAMM)~\citep{TAMM} is a probabilistic graphical model designed for unsupervised discovery of recurrent temporal patterns in time series. It uses non-parametric Bayesian methods fitted using Gibbs sampling to describe both motifs and their temporal occurrences in documents. It is important to mention that the extracted motifs include a temporal dimension. This modeling capability seems to be very interesting in patient phenotyping to derive temporal phenotypes. 
TAMM relies on an improved version of the Probabilistic Latent Sequential Motif model~\citep{VaradarajanBMVC2010} which explains how the set of all observations is supposed to be generated. 
Variable Length Temporal Analysis of Motif Mixtures is a generalization of TAMM that allows motifs to have different lengths and infers the length of each motif automatically. 
The primary limitation is that they do not scale as effectively as the optimization techniques employed in tensor decomposition~\citep{kolda2009tensor}.
Furthermore, the models are highly inflexible, and making modifications requires developing new samplers. 
These limitations prevented us from using these topic models.

\item[Phenotypes as Embeddings]
In the context of neural networks, embeddings are low-dimensional, learned continuous vector representations of discrete variables.  Neural network embeddings are useful because they can reduce the dimensionality of categorical variables and meaningfully represent categories in the transformed space. The primary purposes of using embeddings are finding nearest neighbours in the embedding space and visualizing relations between categories. They can also be used as input to a machine learning model for supervised tasks. \cite{Hettige0LLB20}~introduces MedGraph, a supervised embedding framework for medical visits and diagnosis or medication codes taken from pre-defined standards in healthcare such as International Classification of Diseases (ICD). MedGraph leverages both structural and temporal information to improve the embeddings quality.

\item[Pattern Mining Methods]
Sequential pattern mining~\citep{fournier2017survey} addresses the problem of discovering hidden temporal patterns. A sequential pattern would represent a phenotype by a sequence of events. 
The well-known problem of pattern mining, which tensor decomposition does not suffer from, is pattern deluge. This problem makes it unsuitable for practical use. 
Nonetheless, tensor decomposition methods are close to pattern mining approaches based on compression. GoKrimp~\citep{gokrimp}, SQS~\citep{sqs} and more recently SQUISH~\citep{squish} proposed sequential pattern mining approaches that optimize a Minimum Description Length (MDL) criteria~\citep{galbrun2022minimum}. Unfortunately, only GoKrimp is able to handle sequences of item-sets (\ie with parallel events), but it extracts only sequences of items and does not allow interleaving occurrences of patterns. As representing the parallel events is a crucial aspect of phenotypes, these techniques can not answer the problem of temporal phenotyping.
\end{description}

\section{Temporal Phenotyping: new Problem Formulation}\label{sec:problem}
This section formalizes the problem of temporal phenotyping that is addressed in the remainder of the article. 
In short, temporal phenotyping is a tensor decomposition of a third-order temporal tensor discovering phenotypes that are manifested as temporal patterns. 

\espace

Let $\T{X}$ be an irregular third-order tensor, also viewed as a collection of $K$ matrices of dimension $n \times T_k $, where $K$ is the number of individuals (patients), $n$ is the number of features (care events), and $T_k$ is the duration of the $k$-th individual's observations. 

Given $R\in \mathbb{N}^{*}$, a number of phenotypes and $\omega \in \mathbb{N}^{*}$ the duration of phenotypes (also termed as \textit{phenotype size}), temporal phenotyping aims to build:
\begin{itemize}
    \item $\T{P}\in\mathbb{R}_{+}^{R \times n \times \omega}$: a third-order tensor representing the $R$ temporal phenotypes shared among all individuals. 
    Each temporal phenotype is a matrix of size $n \times \omega$. 
    A phenotype represents the presence of an event at a relative time $\tau$, $0\leq\tau<\omega$. 
    $\omega$ is the same for all phenotypes. 
    $\V{p}_{\tau}^{(r)}$ denotes the vector representing the co-occurring events in the $r$-th phenotype at the relative temporal position $\tau$.
    \item $\T{W}=\left\{\M{W}^{(k)}\in\mathbb{R}_{+}^{R \times T'_{k}} \right\}_{k\in[K]}$: a collection of $K$ assignment matrices of dimension $R \times T'_k$ where $T'_k=T_k-\omega+1$ is the size for the $k$-th individual along the temporal dimension. 
    A non-zero value at position $(r,t)$ in $\M{W}^{(k)}$ describes the start of the phenotype $r$ at time $t$ for the $k$-th individual.
    A matrix $\M{W}^{(k)}$ is also named the \textit{pathway} of the $k$-th individual as it describes his/her history as a sequence of temporal phenotypes.
\end{itemize}

These phenotypes and pathways are built to accurately reconstruct the input tensor, \ie $\T{X}$.
The reconstruction we propose is based on a convolution operator that takes into account the time dimension of $\T{P}$ to reconstruct the input tensor from $\T{P}$ and $\T{W}$.
The convolution operator, denoted $\circledast$, is such that $\M{X}^{(k)} \approx \widehat{\M{X}}^{(k)}=\T{P}\circledast\M{W}^{(k)}$ for all $k\in[K]$ (we remind the reader that $\M{X}^{(k)}$ is the matrix for the $k$-th patient, see section~\ref{sec:soa:tensorbasedrepresentation} and Figure~\ref{fig:tensor}). 
Formally, this operator reconstructs each vector of the matrix $\widehat{\M{X}^{(k)}}$ at time $t$, denoted $\widehat{\V{x}}_{.,t}^{(k)}$, as follows: 
\begin{equation}\label{eq:deconvolution}
\widehat{\V{x}}_{.,t}^{(k)}=\sum_{r=1}^{R}\sum_{\tau=1}^{\min(\omega,t-1)} \V{w}^{(k)}_{r, t-\tau}\V{p}_{\tau}^{(r)}.
\end{equation}

Intuitively, $\widehat{\V{x}}_{.,t}^{(k)}$ is a mixture of phenotype columns that occurred at most $\omega$ time units ago, except at the beginning. 
At one time instant, the observed events are the sum of the $\tau$-th day of the $R$ phenotypes weighted by the $\M{W}^{(k)}$ matrix.

\begin{figure*}[tb]
\centering
\includegraphics[width=.9\linewidth]{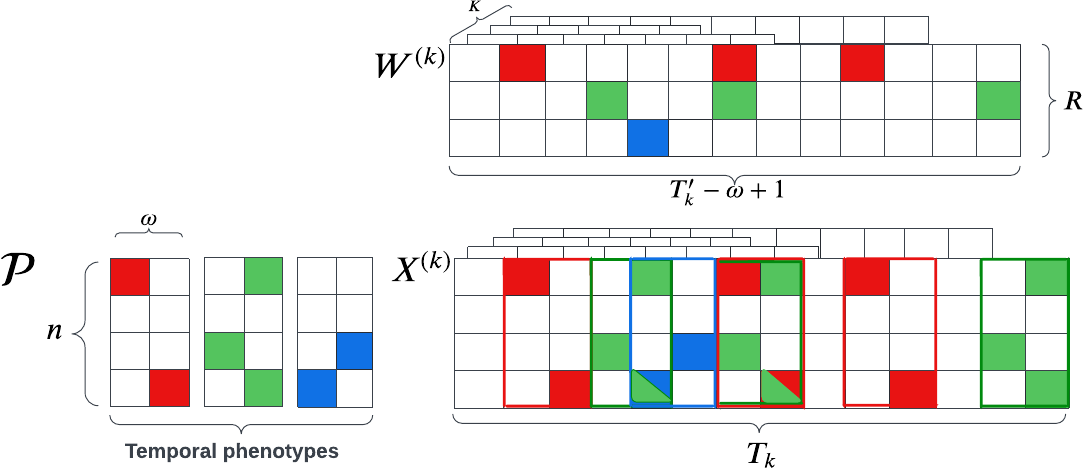}
\caption{Illustration of a matrix reconstruction ($\M{X}^{(k)}$) from $R=3$ phenotypes of size $\omega=2$ on the left and a care pathway ($\M{W}^{(k)}$) on the top. 
Each phenotype has a specific color. 
Each colored cell in $\M{W}^{(k)}$ designates the start of a phenotype occurrence in the reconstruction (surrounded with a colored rectangle in $\M{X}^{(k)}$). A cell with two colors received the contribution of two occurrences of different phenotypes.}
\label{fig:deconvolution}
\end{figure*}

Figure~\ref{fig:deconvolution} depicts the reconstruction of one matrix $\M{X}^{(k)}$ of an input tensor. 
This matrix is of length $T_k=14$ with $n=4$ features. 
Its decomposition is made of $R=3$ phenotypes of size $4\times 2$ each ($\omega=2$ and $n=4$) and a pathway of length $T'_k=14-2+1=13$.  
A colored square in $\M{W}^{(k)}$ indicates the start of phenotype occurrences, which can overlap in $\M{X}^{(k)}$. 
For instance, the column $\V{x}^{(k)}_{.,5}$ combines the occurrence of the second day  of the second phenotype (in green) and the first day of the third phenotype (in blue). 
Each patient is given a pathway matrix $\M{W}^{(k)}$ based on the same phenotypes $\T{P}$ and according to his input matrix $\M{X}^{(k)}$. 
This means that a phenotype represents a typical pattern that might occur in the pathways of multiple patients. 

The problem of temporal phenotyping consists in discovering both ~$\T{P}$ and~$\T{W}$ tensors that reconstruct accurately the input tensor.

\section{\swotted Model} \label{sec:swotted}

\swotted is a tensor decomposition model for temporal phenotyping. 
The generic problem of temporal phenotyping presented above is complemented by some additional hypotheses to guide the solving toward practically interesting solutions. 
These hypotheses are implemented through the definition of a reconstruction loss and regularization terms. This section presents the detail of the \textbf{S}liding \textbf{W}ind\textbf{o}w for \textbf{T}emporal \textbf{Te}nsor \textbf{D}ecomposition model.

\subsection{Temporal Phenotyping as a Minimization Problem}
As in the case of the classic tensor decomposition problem, temporal phenotyping is a problem of minimizing the error between the input tensor and its reconstruction. 

\swotted considers the decomposition of binary tensors, \ie $\T{X}\in\{0,1\}$. 
It corresponds to data that describe the presence/absence of events. 
In this case, we assume the input tensor $\T{X}$ follows a Bernoulli distribution and we use the loss function for binary data proposed by~\cite{Hong_2020}\footnote{\cite{Hong_2020} discuss in detail the choice of the loss function regarding the distribution of a variable: Gaussian, Poisson (positive counts), Gamma (positive continuous data) or Bernoulli (binary data).}.
In the previous section, Equation~\ref{eq:deconvolution} details the reconstruction of a patient matrix. 
The resulting reconstruction loss $\mathcal{L}^{\mathsmaller \circledast}$ is defined as follows:

\begin{equation}\label{eq:SWloss}
\mathcal{L}^{\mathsmaller \circledast}(\hat{\T{X}},\T{X})=\sum_{k=1}^{K} \sum_{t=1}^{T_k} \sum_{i=1}^{n} \log(\hat{x}^{(k)}_{i,t}+1) - x^{(k)}_{i,t} \log(\hat{x}^{(k)}_{i,t}).
\end{equation}
This reconstruction loss is super-scripted by $\circledast$ to remind that it is based on the convolution operator described in Equation~\ref{eq:deconvolution}.

\swotted also includes two regularization terms: sparsity and non-succession regularization. 
Sparsity regularization on $\mathcal{P}$ aims to enforce feature selection and improve the interpretability of phenotypes. It is implemented through an $L_1$ term. We chose this popular regularization technique among several others, as it has shown its practical effectiveness.

\begin{figure}[t!]
\centering
\includegraphics[width=.6\linewidth]{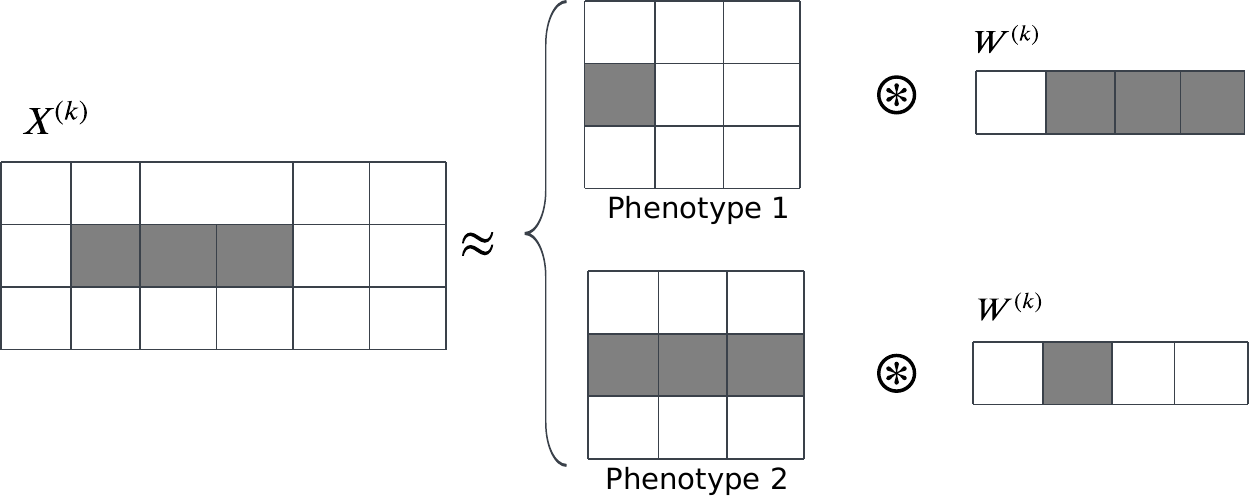}
\caption{Example of alternative decompositions of a sequence of similar events with the same $\mathcal{L}^{\mathsmaller \circledast}$ value. Phenotype~1 does not capture the sequence of events, whereas phenotype~2 does.
The information is reported in the pathway in the case of phenotype~1.}
\label{fig:temp_sliding_tensor_decomp_problem}
\end{figure}

We also propose a phenotype non-succession regularization to prevent undesirable decomposition, as illustrated in Figure~\ref{fig:temp_sliding_tensor_decomp_problem}. 
The described situation is a successive occurrence of the same event. This situation is often encountered in care pathways as a treatment might be a care delivery over several days. 
In this case, there are two opposite alternatives to decompose the matrix with equal reconstruction errors ($\mathcal{L}^{\mathsmaller \circledast}$): the first alternative (at the top) is to describe the treatment as a daily care delivery and to assume that a patient received the same treatment three days in a row; 
the second alternative (at the bottom) is to describe the treatment as a succession of three care deliveries, but that is received only once. 
\swotted implements the second solution as one of our objective is to unveil \textit{temporal} patterns, \ie phenotypes that correlate temporally some events.

To guide the decomposition toward our preferred one, we add a term to penalize a reconstruction that uses the same phenotype on successive days. 
If a phenotype occurs on one day, its recurrence within the following $\omega$ days will incur a cost. 
Formally, the non-succession regularization is defined as follows and depends only on the patient pathway $\M{W^{(k)}}$:

\begin{equation}
\mathcal{S}(\M{W}^{(k)}) = \sum_{r=1}^{R} \sum_{t=1}^{T'_k} \max\left(0, w^{(k)}_{r,t} \log \left( \sum_{\tau=t-\omega}^{t+\omega} w^{(k)}_{r,\tau} \right)\right).
\label{pheno_succ}
\end{equation}
This equation can be seen as a weighted logged convolution where the weight is $w_{r,t}$. Intuitively, as the prevalence of the phenotype grows, the cost of a new occurrence within the same time window also rises.

The inner $\log$ term sums all possible undesirable occurrences of the same phenotype $r$ at time $t$. 
The $\log$ function is used to attenuate the effect of this term and to have a zero value when $w^{(k)}_{r,t}=1$ surrounded with $0$ (the ideal case, depicted in the second decomposition of Figure~\ref{fig:temp_sliding_tensor_decomp_problem}). 

\espace

The final loss function of \swotted is given by the weighted sum of the reconstruction error, the sparsity and the non-succession regularization:

\begin{equation}
 \ell = \mathcal{L}^{\mathsmaller \circledast}\left(\T{P}\circledast\T{W}, \T{X}\right)
 + \alpha\,\, || \T{P} ||_1
 + \beta\, \sum_{k=1}^{K} \mathcal{S}\left(\M{W}^{(k)}\right) 
\label{eq:SW_loss}
\end{equation}
where $\alpha$ and $\beta$ are two positive real-valued hyperparameters. 
Note that the two regularization terms have opposite effects: sparsity encourages phenotypes with many zeros, while non-succession favors the use of non-zero values in phenotypes rather than in pathways. The choice of hyperparameters may impact the quality of the discovered phenotypes.

\subsection{Optimization Framework}
We aim to uncover temporal phenotypes by minimizing the overall loss function $\ell$. The specification of our minimization problem is as follows:

\begin{equation}
\begin{aligned}
& \underset{\{\M{W}^{(k)}\}, \T{P}}{\arg \min}
&  \mathcal{L}^{\mathsmaller \circledast}\left(\T{P}\circledast\T{W}, \T{X}\right)
 + \alpha\,\, || \T{P} ||_1
 + \beta\, \sum_{k=1}^{K} \mathcal{S}\left(\M{W}^{(k)}\right)  \\
& \text{subject to}
& \forall k,\, 0\leq \M{W}^{(k)} \leq 1, \quad 0\leq \T{P} \leq 1.
\end{aligned},
\label{eq:SWOTTED_minimization}
\end{equation}

The minimization problem presented in Equation~\ref{eq:SWOTTED_minimization}  restricts the values' range to $[0,1]$ in order to interpret \T{P} (resp. \T{W}) as the probability of having an event (resp. a phenotype) at a given time. 
These constraints align with our assumption of Bernoulli distribution. 
Another motivation to normalize the pathways is related to the non-succession regularization, $\mathcal{S}(\M{W}^{(k)})$. 
If the values of $\M{W}^{(k)}$ are higher than 1, the non-succession regularization penalizes each presence of a phenotype which is not desired.

\espace

To optimize the overall loss function $\ell$ (see Equation~\ref{eq:SW_loss}), we use an alternating minimization strategy and projected gradient descent (PGD). Alternating Gradient Descent algorithm optimizes one variable at a time, individually, using a gradient descent step, with all other variables fixed. Alternating the process of minimization guarantees reduction of the cost function, until convergence.
PGD handles the non-negativity and the normalization constraints. It works by clipping the values after each iteration.

\espace

\RestyleAlgo{ruled}

\begin{algorithm}[tb]\label{algo:swotted_als}
\DontPrintSemicolon
\caption{Optimization Framework for \swotted}\label{alg:opt-algo}
\KwData{ $\mathcal{X}=\{\M{X^{(k)}}\,\mid\, k \in [K] \}$: patient stays, 
     $R$: the number of temporal phenotypes, 
     $\omega$: the size of the temporal window,
     $\alpha$, $\beta$: loss hyper-parameters.
     }
\KwResult{$\T{P}$: phenotype tensor, 
     $\{\M{W^{(k)}}\,\mid\, k \in [K] \}$: phenotype occurrences in patient stays}
\KwNotation{$\mathrm{clip}(x, 0, 1) = \min(\max(x, 0), 1)$}

$\T{P} \gets \mathcal{U}(0,1)$\tcp*{Random initialization of optimization variables}
$\T{W} \gets \mathcal{U}(0,1)$\;
\For{each epoch}{
     \For{each mini-batch $\mathcal{B}\subset \mathcal{X}$}{
         Update \T{P} by descending along the gradient direction \;
         $ \T{P} \gets \mathrm{clip}(\T{P},0,1) $\; 
    
         \For{$k \in \mathcal{B}$}{
            Update $\M{W}^{(k)}$ by descending along the gradient direction\;
              $ \M{W}^{(k)} \gets \mathrm{clip}(\M{W}^{(k)},0,1) $ \;
         }
     }
 }
\end{algorithm}

The optimization framework of \swotted is illustrated in Algorithm~\ref{alg:opt-algo}. 
$\mathcal{W}$ and $\mathcal{P}$ are initialized with random values drawn from a uniform distribution between 0 and 1. 
In each mini-batch, we first sample a collection of patient matrices ${\{\M{X}^{(k)}\,\mid\,k \in \mathcal{B}\}}$ with $\mathcal{B}$ being the patient's indices of a batch. The phenotype tensor $\T{P}$ is firstly optimized given ${\{\M{W}^{(k)}\,\mid\,k \in \mathcal{B}\}}$ values, then ${\{\M{W}^{(k)}\,\mid\,k \in \mathcal{B}\}}$ is optimized given $\T{P}$ values. 
Note that the gradients are not explicitly computed, but evaluated by automatic differentiation~\citep{baydin2018automatic}. 
The algorithm stops after a fixed number of epochs, with the number of epochs being a predefined hyper-parameter.\footnote{An early stopping based on a convergence criteria would be possible to reduce the computing time of simplest decompositions.}

Among the conventional optimization hyper-parameters, including learning rate, batch size, and the number of epochs, the primary hyperparameters for \swotted encompass $R$ (number of temporal phenotypes), $\omega$ (temporal size of phenotypes), and $\alpha$, $\beta$  (loss weights). In contrast to recent deep neural network architectures, \swotted features a limited set of interpretable hyperparameters.

\subsection{Applying \swotted on Test Sets}\label{sec:projection}
The tensor decomposition presented in the previous section corresponds to the training of \swotted on an irregular tensor $\T{X}$. This training provides a set of temporal phenotypes $\T{P}$. In the minimization problem presented in Equation~\ref{eq:SWOTTED_minimization}, the assignment tensor $\T{W}$ contains a set of free parameters that are to be discovered during the learning procedure but are not kept in the model because they are specific to the train set. 

The results of the decomposition is evaluated on a different test set, $\T{X}'$.
The objective is to assess whether the unveiled phenotypes are useful for decomposing new care pathways. In this case, we can conclude that it captures generalizable phenotypes, otherwise it discovers too specific ones (overfitting). 

Applying a tensor decomposition on a test set, $\T{X}'$, consists in finding the optimal assignment given a fixed set of temporal phenotypes. $\T{X}'$ is a third-order tensor with $K'$ individuals, each having their duration, but sharing the same $n$ features defined in the training dataset. 
The optimal assignment is obtained by solving the following optimization problem that is similar to Equation~\ref{eq:SWOTTED_minimization}, but with a fixed $\widehat{\T{P}}$ (the optimal phenotypes obtained from the decomposition of a train set):

\begin{equation}
\begin{aligned}
& \underset{\T{W}'}{\arg \min}
&  \mathcal{L}^{\mathsmaller \circledast}\left(\widehat{\T{P}}\circledast\T{W}', \T{X}'\right)
 + \beta\, \sum_{k=1}^{K} \mathcal{S}\left(\M{W}'^{(k)}\right)  \\
& \text{subject to}
& 0\leq \T{W} \leq 1.
\end{aligned}
\label{eq:SWOTTED_testset}
\end{equation}
This optimization problem can be solved by a classical gradient descent algorithm. Similarly to the training, we use PGD for the normalization constraint. 

\section{Experimental Setup}\label{sec:setup}

\swotted is implemented in Python using the PyTorch framework~\citep{NEURIPS2019_9015}, along with PyTorch Lightning\footnote{\url{https://www.pytorchlightning.ai}} for easy integration into other deep learning architectures. The model is available in the following repository: \url{https://gitlab.inria.fr/hsebia/swotted}. 
In the experiments, we used two equivalent implementations of \swotted: a classic version that handles irregular tensor and a fast-version that handles only regular tensor benefiting from improved vectorial optimization\footnote{Appendix~\ref{sec:appendix:cmp_fastswotted} provides comparison between the two implementations.}.
Additionally, we provide the repository which includes all the materials needed to reproduce the experiments except for the case study from section~\ref{sec:casestudy}: \url{https://gitlab.inria.fr/tguyet/swotted_experiments}. All experiments have been conducted with desktop computers\footnote{Processor Intel i7-1180G7, 4.60~GHz, 16~Gb RAM, without graphical acceleration, running Ubuntu 22.04 system.}, without the use of GPU acceleration.

We trained the model with an Adam optimizer to update both tensors~\T{P} and~\T{W}. The learning rate is set to~$10^{-3}$ with a batch size of $50$~patients. We fine-tuned the hyperparameters $\alpha$ and~$\beta$ by testing different values and selecting the ones that yielded the best reconstruction measures (see experiments in Section~\ref{sec:results:params}). 
The tensors~\T{P} and~\T{W} are initialized randomly using a uniform distribution ($\mathcal{U}(0,1)$).

The quality measures reported in the results have been computed on test sets. For each experiment, 70\% of the dataset is used for training, and 30\% is used for testing. The test set patients are drawn uniformly.

\subsection{Datasets}

In this section, we present the open-access datasets used for quantitative evaluations of \swotted, including comparisons with competitors. 
We conducted experiments on both synthetic and real-world datasets to evaluate the reconstruction accuracy of \swotted against its competitors. 
Synthetic datasets are used to quantitatively assess the quality of the hidden patterns as they are known in this specific case. 

\subsubsection{Synthetic Data}
The generation of synthetic data involves the reverse process of the decomposition. Generating a dataset follows three steps:
\begin{enumerate}
\item A third-order tensor of phenotypes \T{P} is generated by randomly selecting a subset of medical events for each instant of the temporal window of each phenotype.
\item The patient pathways \T{W} are generated by randomly selecting the days of occurrence for each phenotype along the patient's stay, ensuring that the same phenotype cannot occur on successive days. Bernoulli distributions with $p=0.3$ are used for this purpose. 
\item The patient matrices of \T{X} are then computed using the reconstruction formulation proposed in Equation~\ref{eq:deconvolution}. 
\end{enumerate}
However, this reconstruction can result in values greater than $1$ when multiple occurrences of phenotypes sharing the same drug or procedure deliveries accumulate. To fit our hypothesis of binary tensors, we binarize the tensor resulting from the process above by projecting non-zero values to $1$.

The default characteristics of the synthetic datasets subsequently generated for various experiments are as follows: $K=500$ patients, $n=20$ care events, $R=4$ phenotypes of length $\omega=3$, and stays of $T_k=10$ days for all $k$.

\subsubsection{Real-world Datasets} 
The experiments conducted on synthetic datasets are complemented by experiments on three real-world datasets, which are publicly accessible. 
We selected one classical dataset in the field of patient phenotyping (namely the MIMIC database) and two sequential datasets coming from very different contexts. Table~\ref{tab:datasets} summarizes the main characteristics of these datasets. 

\begin{itemize}
\item MIMIC dataset\footnote{Original dataset: \url{https://physionet.org/content/mimiciv/0.4/}}: MIMIC-IV is a large-scale, open-source and deidentified database providing critical care data for over~$40,000$ patients admitted to intensive care units at the Beth Israel Deaconess Medical Center (BIDMC)~\citep{johnson2020mimic}.  We used the version 0.4 of MIMIC-IV.
The dataset has been created from the database by selecting a collection of patients and gathering their medical events during their stay. 
For the sake of reproducibility, the constitution of the dataset is detailed in Appendix~\ref{sec:expesetup}. In addition, the code used to generate our final dataset is provided in the repository of experiments.

\item E-Shop dataset\footnote{Original data: \url{https://archive.ics.uci.edu/dataset/553/clickstream+data+for+online+shopping}. We used the prepared version from \href{https://www.philippe-fournier-viger.com/spmf/index.php?link=datasets.php\#r1}{SPMF repository}.}: This  dataset contains information on clickstream from one online store offering clothing for pregnant women. Data are from five months of 2008 and include, among others, product category, location of the photo on the page, country origin of the IP address and product price. 

\item Bike dataset\footnote{Original data: \url{https://www.kaggle.com/cityofLA/los-angeles-metro-bike-share-trip-data}. We used the prepared version from \href{https://www.philippe-fournier-viger.com/spmf/index.php?link=datasets.php\#r1}{SPMF repository}. }: This contains sequences of locations where shared bikes where parked in a city. Each item represents a bike sharing station and each sequence indicate the different locations of a bike over time.
The specificity of this dataset is to contain only one location per date. 
\end{itemize}

For each of these datasets, we also created a ``regular'' version, which contains individuals' pathways sharing the same length. This dataset is utilized with our fast-\swotted implementation that benefits from better vectorization. The creation of these datasets involves two steps: 1) selecting individuals with pathway durations greater than or equal to $T$, and 2) truncating the end of the pathways if they exceed $T$ in length. The selection of the values for $T$ is a balance between maintaining the maximum length and retaining the maximum number of individuals.

\begin{table}[tbp]
\caption{Real-world dataset characteristics: $n$ number of features, $K$ number of individuals, $\bar{T}$ mean duration}\label{tab:datasets}
\begin{tabular}{lccc}
\toprule
Dataset & $n$ & $K$ & $\bar{T}$ \\
\midrule
MIMIC-IV & $2,717$ & $200$ & $9.62\pm 12.7$\\ 
MIMIC-IV-reg & $825$ & $200$ & $10$ \\
Bike & $21,078$& $67$& $11.43\pm 5.93$ \\
Bike-reg & $5,243$ & $67$& $10$ \\
E-Shop & $24,026$ & $317$ & $13.79\pm 11.19$\\ 
E-Shop-reg & $5,212$ & $317$ & $10$\\
\bottomrule
\end{tabular}
\end{table}

We remind that our case study presents another dataset of patients staying in the Greater Paris University Hospitals for qualitative analysis of phenotypes. This dataset will be detailed in Section~\ref{sec:casestudy}.

\subsection{Competitors}\label{sec:competitors}
We compare the performance of \swotted against four state-of-the-art tensor decompostion models. 
These models were selected based on the following criteria: 1) their motivation to analyze EHR datasets, 2) their competitiveness in terms of accuracy compared to other approaches, 3) the availability of their implementations, and 4) their handling of temporality. 

We remind that \swotted is the only tensor decomposition technique able to extract temporal patterns. Our competitors extract daily phenotypes.

The four competing models are the followings:
\begin{itemize}
\item \textit{CNTF}~--~Collective Non-negative Tensor Factorization~\citep{CNTF_2019} a tensor decomposition model factorizing tensors with varying temporal size, assuming the input tensor to follow a Poisson distribution, but it has shown its effectiveness on binary data; CNTF is our primary competitor since it incorporates temporal regularization, aiming to capture data dynamics. 
\item \textit{PARAFAC2}~\citep{kiers1999parafac2}, original decomposition model with non-negative constraint. This decomposition is based on Frobenius norm. 
We use the Tensorly implementation~\citep{kossaifi2016tensorly}.
\item \textit{LogPar}~\citep{yin2020logpar}, a logistic PARAFAC2 for learning low-rank decomposition with temporal smoothness regularization. We choose to include LogPar in the competitors' list because, like \swotted, it is designed for binary tensors and assumes a Bernoulli distribution. LogPar can handle only regular tensors.

\item \textit{SWIFT}~--~Scalable Wasserstein Factorization for sparse non-negative Tensors~\citep{afshar2020swift}, a tensor decomposition model minimizing the Wasserstein distance between the input tensor and its reconstruction. SWIFT does not assume any explicit distribution, thus it can model complicated and unknown distributions.

\end{itemize}

For each experiment, we manually configure their hyper-parameters to ensure the fairest possible comparisons. 

\subsection{Evaluation Metrics}
In tensor decomposition, a primary objective is to reconstruct accurately an input tensor. 
We adopt the $FIT \in (- \infty, 1]$~\citep{FIT} to measure the quality of a model's reconstruction:

\begin{equation}
 FIT_X = 1 - \frac{\sum_{k=1}^{K} || \M{X}^{(k)} - \M{\widehat{X}}^{(k)}||_F}{\sum_{k=1}^{K} || \M{X}^{(k)} ||_F}
\label{eq:FIT_measure}
\end{equation}
where the input tensor \T{X} serves as the ground truth, the resulting tensor is denoted $\T{\widehat{X}}$ and $||\cdot||_F$ is the Frobenius norm. 
The higher the value of $FIT$, the better. 
The $FIT$ measure is also used to compare phenotypes and patient pathways when hidden patterns are known a priori, \ie, for synthetic datasets. Thus, $FIT_P$ (resp. $FIT_W$) denotes the reconstruction quality of \T{P} (resp. \T{W}).

It is worth noting that $FIT_X$ measure is computed on a test set except for SWIFT and PARAFAC2. Evaluation on test sets requires the model to be able to project a test set on existing phenotypes (see Section~\ref{sec:projection}), but SWIFT and PARAFAC2 do not have this capability.

\espace

We also introduce a similarity measure between two sets of phenotypes to evaluate empirically the uniqueness of solutions and a diversity measure of a set of phenotypes, adapted from similarity measures introduced by~\cite{CNTF_2019}.

Let $\mathcal{P}=\{\M{P}_1,\dots,\M{P}_R\}$ and $\mathcal{P}'=\{\M{P}'_1,\dots, \M{P}'_R\}$ be two sets of phenotypes defined over a temporal window size $\omega$. The principle of our similarity measure is to find the optimal matching between the phenotypes of the two sets, and to compute the mean of the (dis)similarities between the matching pairs of phenotypes. 
More formally, in the case of cosine similarity, we compute: 

$$\mathrm{sim}(\mathcal{P},\mathcal{P}')=\max_{\pi} \left(\begin{array}{c}R-1\\2\end{array}\right)^{-1}\sum_{(i,j)\in\pi} \cos(\M{P}_i,\M{P}'_j)$$

\noindent where $\pi$ denotes an isomorphism between $\mathcal{P}$ and $\mathcal{P}'$, and $\cos(\cdot,\cdot)$ is a cosine distance between two temporal phenotypes. It is computed as the mean of the cosine similarity between each time slice of the phenotype:

$$\cos(\M{P},\M{P}')=\frac{1}{\omega}\sum_{i=1}^{\omega}\frac{ \langle\V{p}_{:,i},\V{p}'_{:,i}\rangle}{||\V{p}_{:,i}||\, ||\V{p}'_{:,i}||}$$

\noindent where $\langle \cdot,\cdot\rangle$ is the Euclidean inner product.

In practice, we first compute a matrix of costs and use the Hungarian algorithm~\citep{kuhn1955hungarian} to find the optimal matching ($\pi$). Finally, we compute the measure with $\pi$.

\espace

The diversity measure of the set of phenotypes aims to quantify the redundancy among the phenotypes. In this case, we expect to have low similarities between the phenotype. 

Let $\mathcal{P}=\{\M{P}_1,\dots,\M{P}_R\}$ be a set of $R$ phenotypes defined over a temporal window of size $\omega$, the cosine diversity is defined by:

$$\mathrm{div}(\mathcal{P})=\left(\frac{R}{2}\right)^{-1}\sum_{1\leq i<j \leq R} 1-\cos(\M{P}_i,\M{P}_j).$$

\section{Experiments and Results}\label{sec:experiments}

In this section, we present the experiments we conducted to evaluate \swotted.

\subsection{Loss Hyper-Parameters Exploration and Ablation Study}\label{sec:results:params}

This section focuses on the loss hyper-parameters of \swotted\footnote{Additional results on the effect of the normalization are presented in Annex~\ref{sec:annex:normalization}.}. The objectives of these experiments are to evaluate the usefulness of each term, to provide a comprehensive review of the effect of the loss parameters and to assess whether the model behaves as expected. 
We start by presenting some experiments on synthetic datasets and then confirm the results on the three real-world datasets.

\begin{sidewaysfigure}[htbp]
\centering
\includegraphics[width=\textwidth]{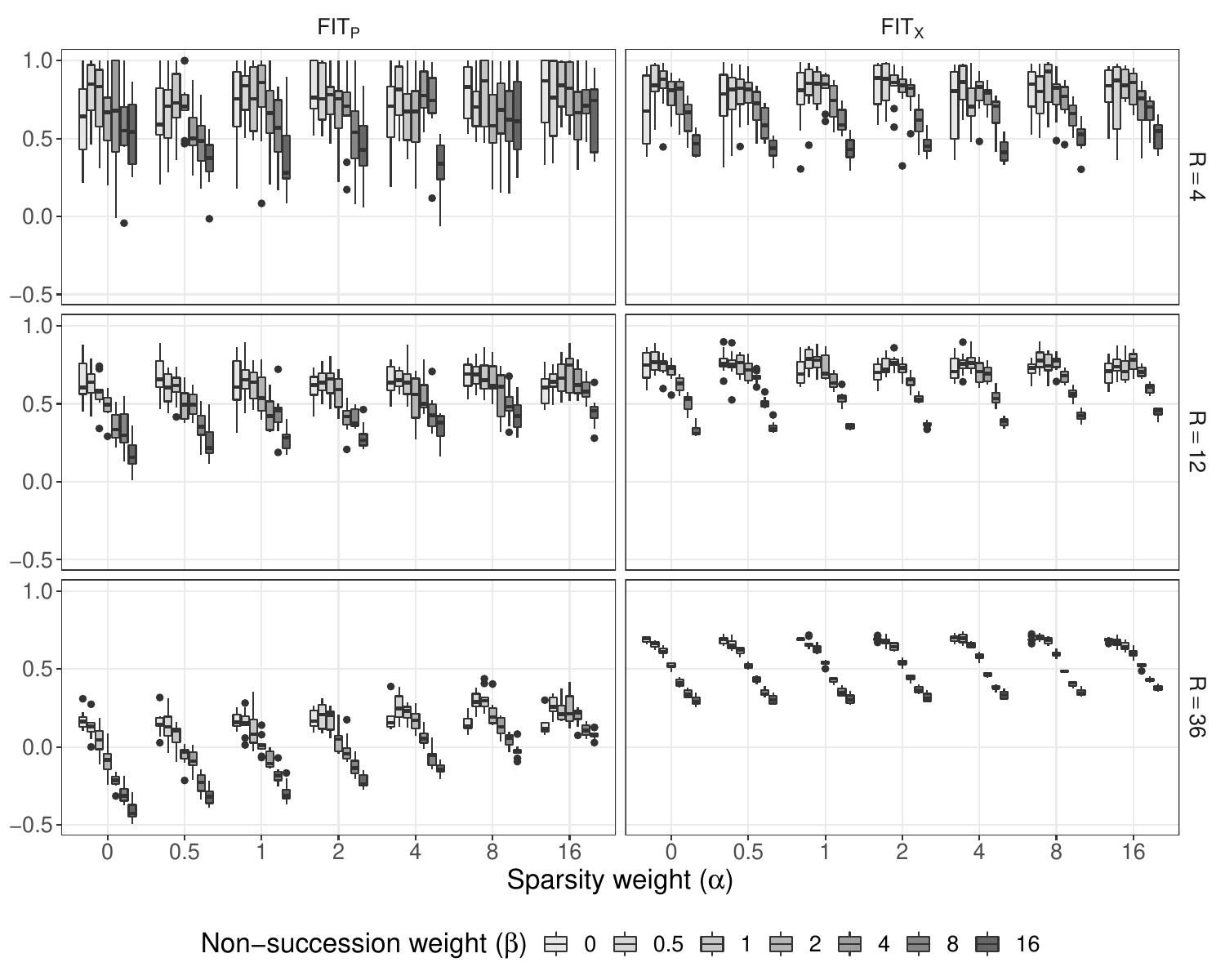}
\caption{$FIT_P$ (left) and $FIT_X$ (right) of \swotted with $\omega=3$ on synthetic data. Each graph represents a box plot for 10 runs. }
\label{fig:result_synth_params}
\end{sidewaysfigure}

\espace

In this first experiment, we generated synthetic datasets using $R$ hidden phenotypes ($R=4, 12$ or $36$) with a window size $\omega=3$. \swotted is run 10 times with different parameter values:
\begin{itemize}
\item $\alpha=\{0, 0.5, 1, 2, 4, 8, 16\}$ representing the weight of the sparsity term in the loss; $\alpha=0$ disables this term.
\item $\beta=\{0, 0.5, 1, 2, 4, 8, 16\}$ representing the weight of the phenotype non-succession term in the loss; $\beta=0$ disables this term.
\end{itemize}

We collected the $FIT_X$ and $FIT_P$ metric values on a test set. The first metric assesses the quality of the reconstruction, while the second assesses whether the discovered phenotypes match the expected ones.

Figure~\ref{fig:result_synth_params} depicts the $FIT$ measures with respect to the parameters $\alpha$ and $\beta$. The normalization is considered in this experiment. 
One general result is that the $FIT_X$ values are high. 
Values exceeding $0.5$ indicate significantly good reconstructions, and those surpassing 0.8 imply that the differences between two matrices become imperceptible. 
Furthermore, we observe that a good $FIT_P$ implies a good $FIT_X$. This illustrates that in tensor decomposition, an accurate discovery of hidden phenotypes is beneficial for achieving a high-quality tensor reconstruction. However, as we see with $R=32$, a good $FIT_X$ does not necessarily mean that the method discovered the correct phenotypes.

A second general observation is that the same evolution of the FIT with respect to~$\beta$ is observed in most of the settings: the $FIT$ increases between $\beta=0$ and $\beta=1$ and then it decreases quickly for higher values. This demonstrates two key points: 1)~the inclusion of non-succession term improves the reconstruction accuracy, and 2)~on average, a value of $\beta=0.5$ yields the best results.
Regarding the parameter~$\alpha$, we notice a slight improvement in~$FIT$ measures as~$\alpha$ increases. This is more obvious with~$R=36$. When we focus on results with $\beta=0.5$, we observe that $\alpha=1$ is, on average, the best compromise for the sparsity term.

One last observation is that as $R$~increases, $FIT$~measures decrease. This may be counter-intuitive, as the expected results is that the $FIT_X$~decreases with increasing~$R$ (as we will discuss with real-world dataset results later on). 
In this experiment, $R$~corresponds to both the rank of the decomposition and the number of hidden phenotypes we used to generate the dataset. 
As $R$~increases, the dataset contains more variability and denser events, making the reconstruction task challenging and leading to a slightly decrease of the $FIT$~measure. $FIT_X$~maintains a high value even with~$R=36$ but~$FIT_P$ has low values in this case. We explain this observation by colinearities between phenotypes.

\espace

Finally, we complemented our analysis of parameters by specifically investigating the non-succession term introduced in \swotted. 
To assess its efficiency, we generated synthetic datasets with~$4$ random phenotypes and $6$ phenotypes that have been designed to contain successions of similar events (see Figure~\ref{fig:temp_sliding_tensor_decomp_problem}). 

\begin{figure}
\centering
\includegraphics[width=0.45\textwidth]{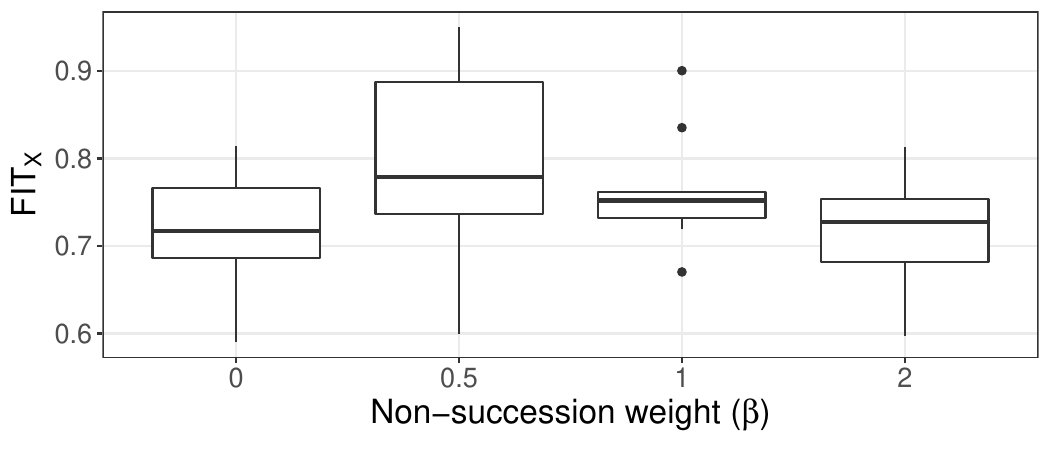}
\includegraphics[width=0.45\textwidth]{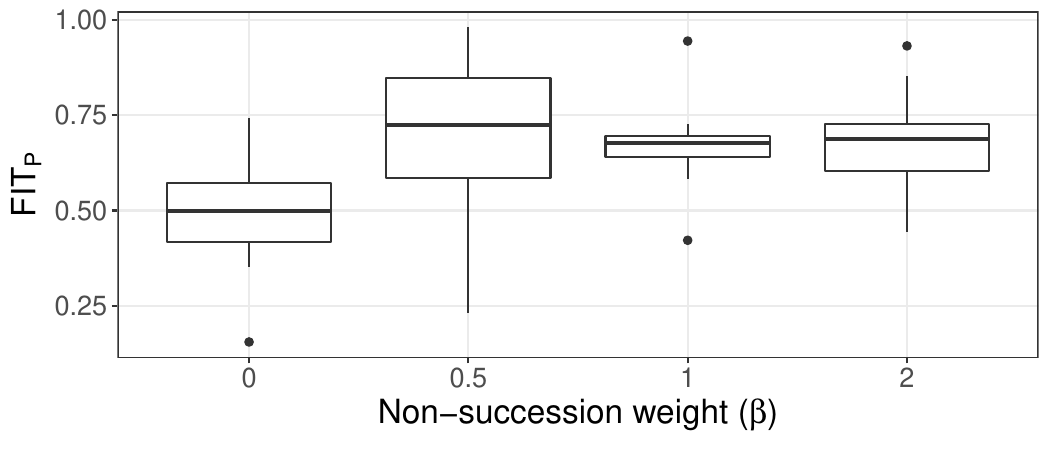}
\caption{Comparison of~$FIT_X$ (left) and~$FIT_P$ (right) with respect to the~$\beta$ hyper-parameter on a synthetic dataset with hidden phenotypes having repeated successive events.}
\label{fig:result_synth_FIT_nonsuccession}
\end{figure}

Figure~\ref{fig:result_synth_FIT_nonsuccession} depicts the~$FIT_X$ and~$FIT_p$ values with respect to~$\beta$ ($\alpha$ is set to $2$ in this experiment). We observe clearly that $FIT_P$ are higher when $\beta$ is not zero, \ie when we use the non-succession term in \swotted ($FIT_P=0.75$ instead of $0.50$ when $\beta=0$). 
The best median $FIT_X$ is close to $0.8$ and occurs when $\beta=0.5$. 
This confirms that adding the non-succession regularization disambiguates the situation illustrated in Figure~\ref{fig:temp_sliding_tensor_decomp_problem} and helps the model to correctly reconstruct the latent variables. 
We conclude that the use of the non-succession regularization increases the decomposition quality, whether there are event repetitions in phenotypes or not (see Figure~\ref{fig:result_synth_params}).

\subsection{\swotted against Competitors}\label{sec:results:competitors}

In this section, we compare \swotted against competitors based on the ability to achieve accurate reconstructions, to extract hidden phenotype effectively and to efficiently handle large scale datasets. To address these various dimensions, we use both synthetic and real-world datasets. 

For the real-world datasets, we use truncated versions because LogPar requires regular tensors. Additionally, we remind that $FIT$ values are evaluated on test sets, except for SWIFT and PARAFAC2, for which we utilize the error on the training set since these approaches can not be applied on test sets.

\subsubsection{\swotted Accuracy on Daily Phenotypes}
This experiment compares the accuracy of \swotted against competitors on 20 synthetic datasets. 
For the sake of fairness, the datasets are generated with daily hidden phenotypes ($\omega=1$). 
Our goal is to evaluate the accuracy of phenotypes extracted by \swotted compared to the ones of state-of-the-art models. 

\begin{figure}[tb]
    \centering
	\includegraphics[width=\columnwidth]{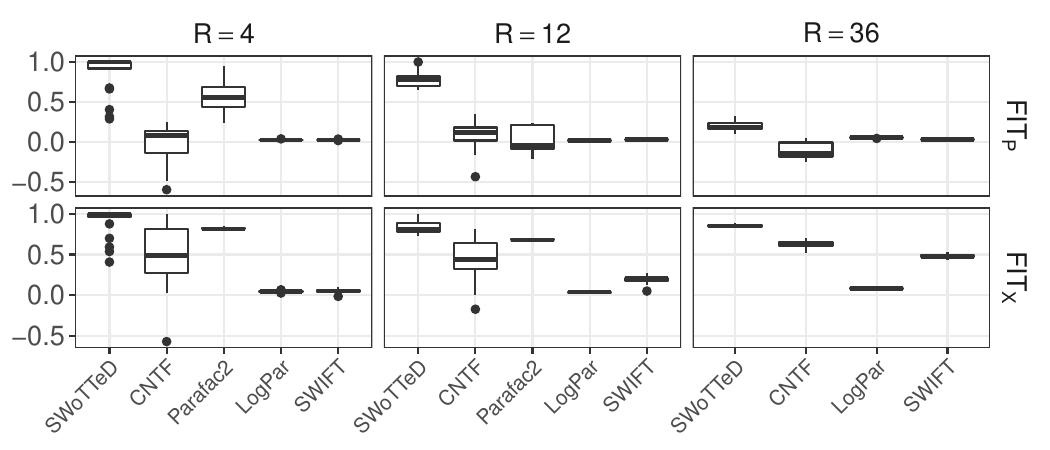}
    \caption{$FIT_P$ values (at the top) and $FIT_X$ values (at the bottom) of \swotted vs competitors on synthetic data with $\omega=1$.}
    \label{fig:result_synth_cmp}
\end{figure}

\begin{figure}[tb]
    \centering
	\includegraphics[width=.49\columnwidth]{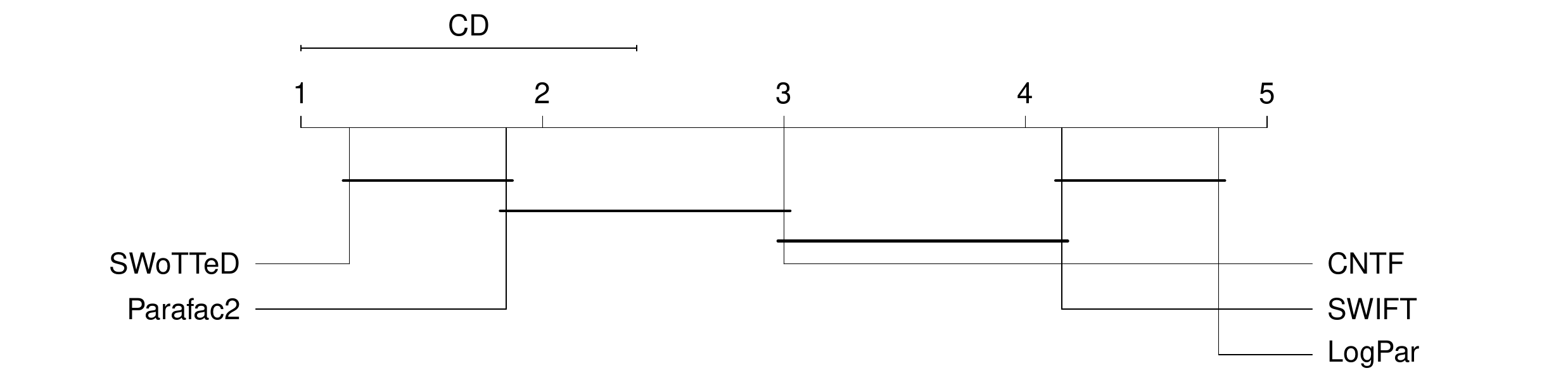}
	\includegraphics[width=.49\columnwidth]{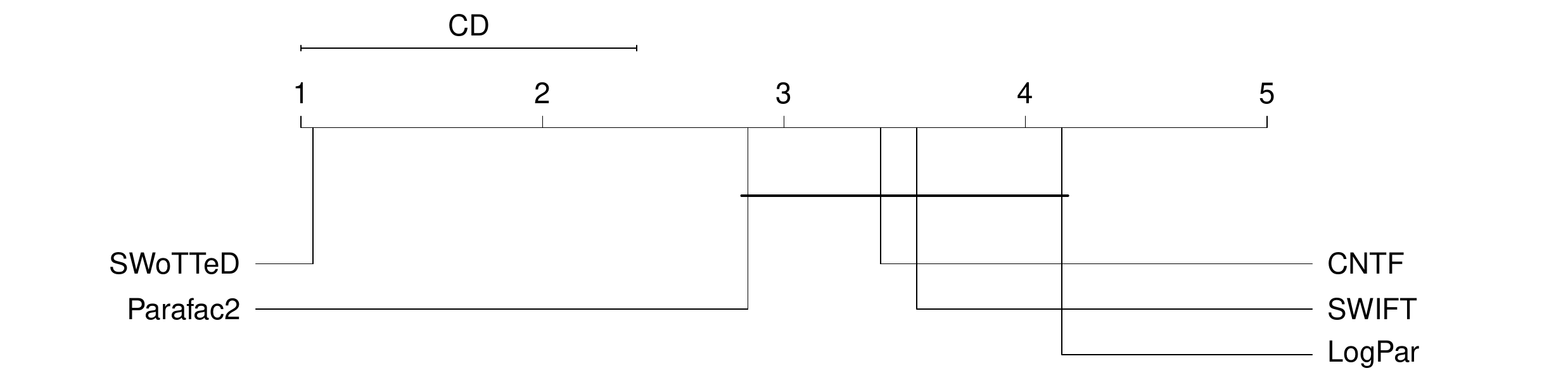}
    \caption{Critical difference diagrams between \swotted and its competitors (based on $FIT_X$ metric on the left, and on $FIT_P$ metric on the right). The lower the rank, the better.  Horizontal bars indicate statistical non-significant difference between models. The significancy is obtained with a Wilcoxon signed-rank test ($\alpha=0.05$).}
    \label{fig:result_synth_cmp_CD}
\end{figure}

The results, depicted in Figure~\ref{fig:result_synth_cmp}, show that \swotted achieves the best performance in terms of $FIT_X$ and $FIT_P$ metrics. 
The second-best model is PARAFAC2. It achieves good tensor reconstructions but it fails to identify the hidden phenotypes (low $FIT_P$ values). The analysis of the phenotypes shows that PARAFAC creates mixtures of phenotypes. Note that PARAFAC2 has no value for $R=36$ because $R$ must be lower than the maximum of every dimension.

The reconstructions of CNTF are also satisfying but the phenotypes are different from the expected one. Our intuition is that assuming Poisson distribution is not effective for these data following a Bernoulli distribution. 
The two other competitors are assumed to be adapted to these data but in practice, we observe that they have reconstructed the tensors with lower accuracy, and the extracted phenotypes are less similar to the hidden ones compared to \swotted.

To confirm these results, Figure~\ref{fig:result_synth_cmp_CD} depicts critical difference diagrams\footnote{Critical diagram \citep{demvsar2006statistical} is a visual representation of the results of the Wilcoxon signed-rank test (see figure legend for reading instructions).}. It ranks the methods based on the Wilcoxon signed-rank test on $FIT_X$ or $FIT_P$ metrics. The diagram shows that \swotted is ranked first, and the difference in $FIT_P$ compared to other methods is statistically significant.

The high-quality performance of \swotted can be attributed to two main factors. Firstly, \swotted offers greater flexibility in reconstructing input data by allowing the overlap of different phenotypes and their arrival with a time lag. Secondly, \swotted employs a loss function that assumes a Bernoulli distribution, which fits better binary data.

\subsubsection{\swotted against Competitors on Real-world Datasets}
In this section, we evaluate \swotted against its competitors on the three real-world datasets to confirm that previous results applies on real-world data. 
We vary~$R$ from~$4$ to~$36$, $\omega$ from~$1$ to~$5$ for \swotted, and we compare $FIT_X$ values. 
It is worth noting that $FIT_P$ can not be evaluated in this case as the hidden phenotypes are unknown. 
Each setting is ran 10 times with different train and test sets.

\begin{figure}
\centering
\includegraphics[width=\textwidth]{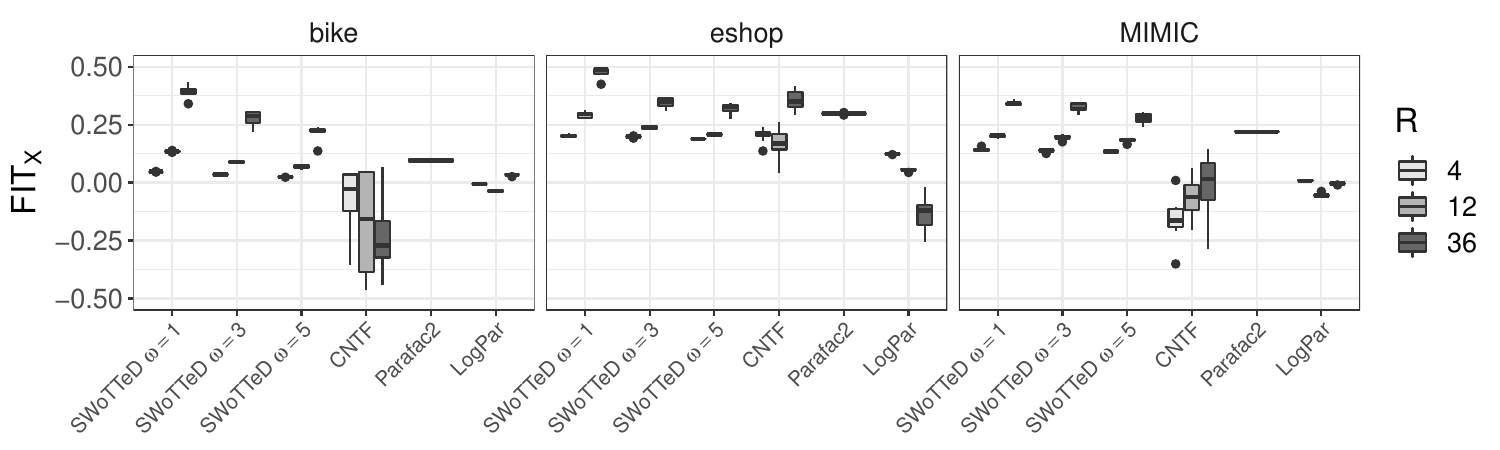}
\caption{Reconstruction error $FIT_X$ of \swotted ($\omega=1, 3, 5$) and its competitors on real-world datasets and for different values of $R$. }
\label{fig:result_RWD_FITX}
\end{figure}

Figure~\ref{fig:result_RWD_FITX} summarizes the results. 
Across all datasets, \swotted achieves an average $FIT_X$ of $0.21$. PARAFAC2 achieves the second best reconstructions but we remind you that $FIT_X$ is evaluated on the train set.  CNTF has good results except for the bike dataset. 
Specifically, \swotted outperforms all the competitors with comparable $R$ on bike and MIMIC datasets regardless of the phenotype size. 
On the E-shop dataset, CNTF exhibits good performances with $R=36$ and outperforms \swotted when $\omega>1$. Nonetheless, \swotted with $\omega=3$ becomes better than CNTF. 
LogPar and SWIFT have the worse $FIT_X$ on average. For the sake of graphic clarity, SWIFT is not represented in this figure. Its $FIT_X$ values are below $-0.5$. The complete Figure is provided in Appendix (see Figure~\ref{fig:result_RWD_FITX_full}, page~\pageref{fig:result_RWD_FITX_full}).

The figure presents results for three different values of $R$. As expected,  $FIT_X$ of \swotted increases with $R$. 
Intuitively, real-world datasets contain a diversity and a large number of hidden profiles. 
With more phenotypes, the model becomes more flexible and can capture the diversity in the data more effectively, resulting in more accurate tensor reconstructions.

We also observe that while $FIT_X$ decreases slightly as the phenotype size~($\omega$) increases, all values remain higher than those of CNTF, except for bike with $R=36$. This suggests that \swotted with $\omega>1$ discovers phenotypes that are both complex and accurate. 
It may seem counter-intuitive that the $FIT$ does not decrease. 
In fact, the number of model parameters increases with the size of the phenotype.
However, these parameters are not completely free. Larger phenotypes also add more constraints to the reconstruction due to temporal relations, which can introduce errors when a phenotype is only partially identified. The more complex is the phenotype, the more likely there is a difference between the mean description and its instances.

The most important result of this experiment is that the reconstruction with temporal phenotypes competes with the reconstruction with daily phenotypes. This means that \swotted strikes a balance between a good reconstruction of the input data and an extraction of rich phenotypes. 
Furthermore, the temporal phenotypes --~with~$\omega$ strictly higher than~$1$~-- convey a rich information to users by describing complex temporal arrangements of events. 

\subsubsection{Time Efficiency}\label{sec:results:time}

\begin{figure}
\centering
\includegraphics[width=\textwidth]{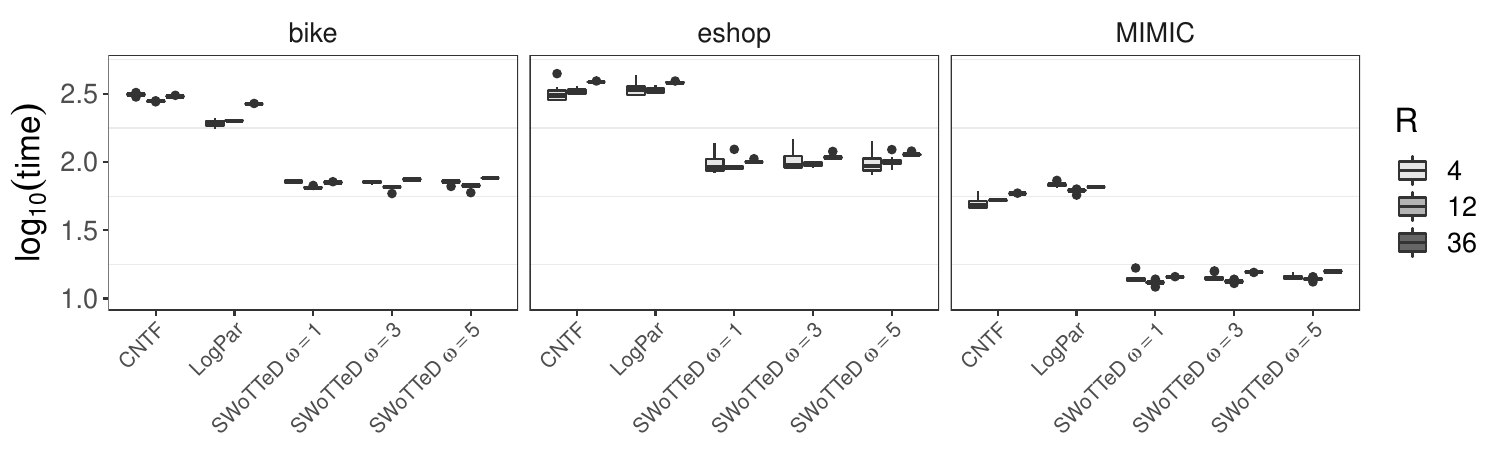}
\caption{Computing time in seconds (base-10 log-scale) of one run of \swotted and its competitors.}
\label{fig:result_RWD_time}
\end{figure}

Figure~\ref{fig:result_RWD_time} illustrates the computing times of the training process on real-world datasets. It compares the computing time of \swotted with its competitors under different setting (varying values of~$R$ and~$\omega$). 
We have excluded SWIFT from this figure as its computing time is several orders of magnitude slower than the other competitors due to the computation of Wasserstein distances. 

This figure shows that our implementation of \swotted is one order of magnitude faster than CNTF or LogPar. 
This efficiency is attained through a vectorized implementation of the model.
The mean computing times on a standard desktop computer\footnote{Intel i7-1180G7, 4.60~GHz, 16~Gb RAM, without graphical acceleration.} for \swotted are $70.42s\pm 18.37$, $102.83s\pm 209.34$ and $14.34s\pm 1.03$ for the Bike, E-shop, and MIMIC datasets respectively.

Regarding the parameters of \swotted, we observe that the computing time grows linearly with the number of phenotypes. More surprisingly, the size of the phenotypes has only a minor impact on computing time. This can be attributed to the efficient implementation of convolution in the PyTorch framework.

Despite the relatively high theoretical complexity of the reconstruction procedure (see Annex \ref{sec:app:complexity}), this experiment demonstrates that \swotted has low computing times and can scale to handle large datasets.


\subsection{\swotted Robustness to Data Noise}\label{sec:results:robustness}
Noisy data are a common challenge encountered in the analysis of medical data. Physicians may make errors during data collection. Some exams may not be recorded in electronic health records and the data collection instruments themselves may be unreliable, resulting in inaccuracies within datasets. These inaccuracies are commonly referred to as noise. Noise can introduce complications as machine learning algorithms may interpret it as a valid pattern and attempt to generalize from it. Therefore, we conducted an assessment of the robustness of our model against simulated noisy data.

We considered two types of common noise that are due to data entry errors: 1) the additive noise due to occurrences of additional events in patient's hospital stays, and 2) the destructive noise, due to important events that have not been reported. 
We start by generating 20~synthetic datasets that are divided into training and test sets. Only the training set is disturbed, and the $FIT$ value is measured over the test set.  
The idea behind disturbing only the training set is to assess \swotted's ability to capture meaningful phenotypes in the presence of noise that can be generalized over a non-disturbed test set.
For additive noise, we inject additional events positioned randomly into the \T{X} tensor. The number of added events per patient is determined according to a Poisson distribution with a parameter $\lambda$. We vary $\lambda$ from $2$ to $25$ with a stepsize of $5$, except for the first step that has a value of 3\footnote{This mean that the values of $\lambda$ are successively 3, 5, 10, 15, 20 and 25.}. The noise level is normalized by the number of ones in the dataset (\ie the number of events). For instance, $\lambda = 0.3$ means that $30\%$ of additional events have been injected into the data. Values greater than $1.0$ for noise addition indicates than more than half of the events are random.
For the destructive noise, we iterate over all the events of all patients in \T{X}, and delete them based on a Bernoulli distribution with a parameter~$p$. We vary $p$ from~$0$ to~$0.7$ with a stepsize of $0.1$. 

\begin{figure}[tb]
    \centering
    \includegraphics[width=1\columnwidth]{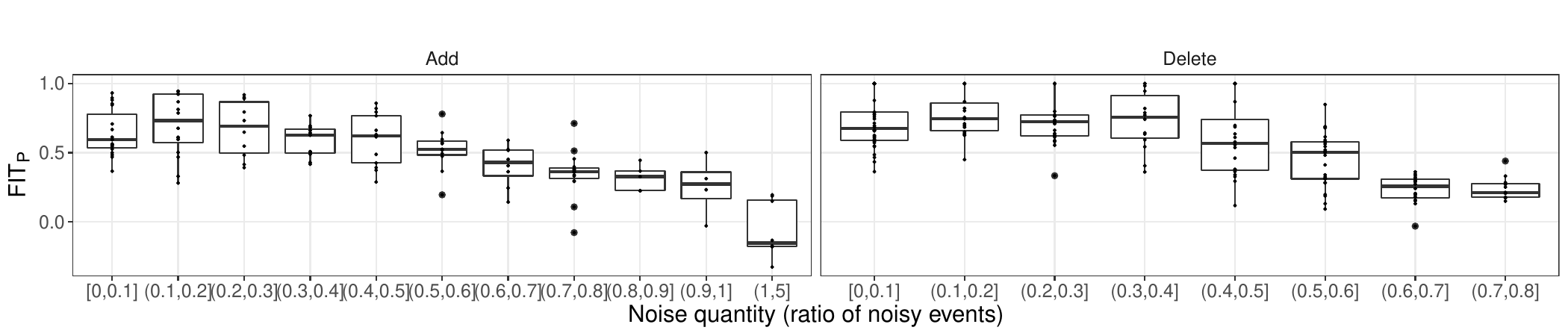}
    \caption{$FIT_P$ values of \swotted on synthetic data as a function of normalized noise (\% of new or deleted events).}
    \label{fig:robustness}
\end{figure}

Our focus was primarily on \swotted's ability to derive correct phenotypes from noisy data, as measured by the $FIT_P$ metric.

Figure~\ref{fig:robustness} displays the values of $FIT_P$ obtained with various noise ratios. 
In the case of added events, we notice that $FIT_P$ decreases as the average number of added events per patient increases. However, the quality of reconstruction remains above zero even when the average number of added events per patient reaches 10. 
In the case of deleting events, we notice that $FIT_P$ starts to decrease when the ratio of missing events exceeds $0.3$. In an extreme case where we have 70\% of missing events, \swotted still manages to have a positive phenotype reconstruction quality.

Consequently, we can conclude that our model exhibits robustness to noisy data, particularly in the case of missing data. 
 This experiment further confirms the interest of tensor decomposition when data are noisy. Interestingly, adding some random noise even resulted in improved accuracy. 
We explain this by the relatively low number of epochs (200): some randomness in the data fasten the convergence of optimization algorithms. With fewer epochs, the model discovered better phenotypes in the presence of noise. 
Being robust to destructive noise is more promising. In real-world dataset, especially in care pathways that is our primary application, missing events might be numerous. The results show that our model discovers the hidden phenotypes with high accuracy even with a lot of (random) missing events.

\subsection{Uniqueness and Diversity in \swotted Results}\label{sec:results:uniqueness}
Solving tensor decomposition problems with an alternating optimization algorithm does not guarantee a convergence toward a global minimum or even a stationary point, but only to a solution where the objective function stops decreasing~\citep{kolda2009tensor}. The final solution can also be highly dependent on the initialization and of the training set. 
Similarly, \swotted does not come with convergence guarantees, but we can empirically evaluate the diversity of solutions obtained across different runs.

The experiments conducted on synthetic datasets illustrated that different runs of \swotted converge toward the expected phenotypes (as detailed in Section \ref{sec:results:params}). 
However, it can not conclusively determine the uniqueness of solutions, as it heavily relies on the random phenotypes that have been generated. We exclusively employ real-world datasets in this section.

\begin{figure}[tb]
    \centering
    \includegraphics[width=\linewidth]{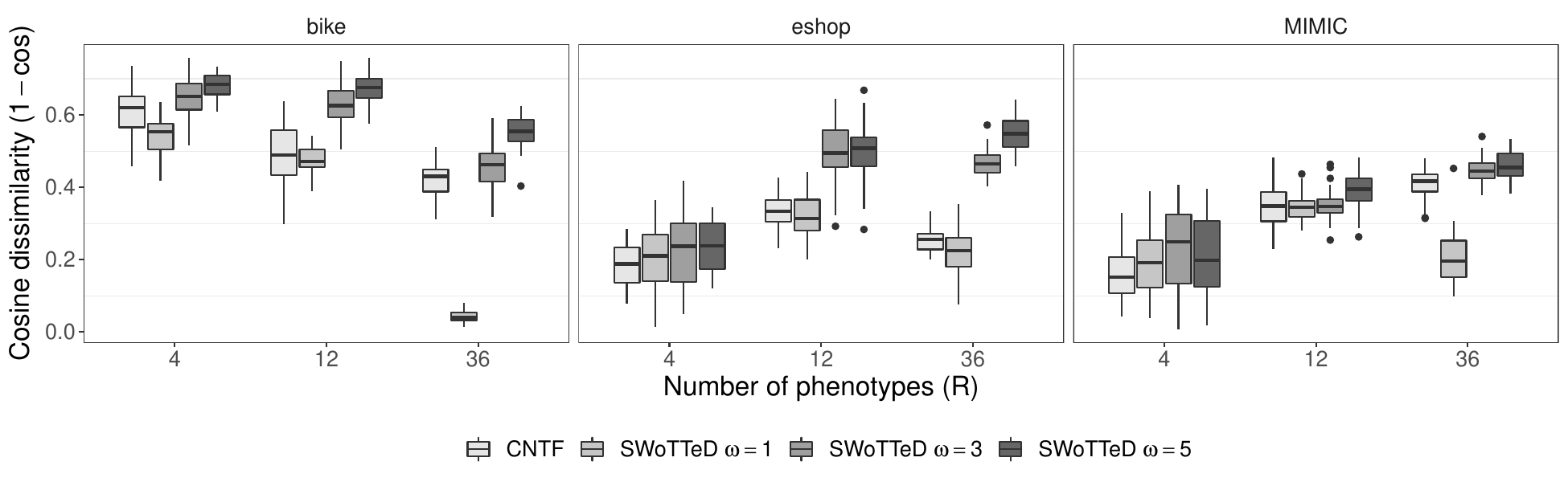}
    \caption{Cosine dissimilarity between pairs of sets of phenotypes with respect to the phenotypes' size for \swotted, and CNTF. The lower the dissimilarity, the more similar are the sets of phenotypes between each run.}
    \label{fig:exp_uniqueness}
\end{figure}

In this experiment, we delve into the sets of phenotypes in our real-world datasets.  
For each dataset, we run \swotted 10~times and compare the sets of phenotypes using average cosine dissimilarity. 

Figure~\ref{fig:exp_uniqueness} depicts the cosine dissimilarity obtained with $R=4$, $12$ and $36$ for \swotted (with varying phenotype sizes) and CNTF.
Lower dissimilarity values indicate greater similarity between phenotypes from one run to another, which is preferable.

With $R=36$, the cosine dissimilarity is below $0.5$ for all datasets.  
In the case of \swotted, it generally exceeds $0.3$. 
This observation suggests that there may be multiple local optima, where the optimization procedure must make choices among the phenotypes to represent. Consequently, the convergence location may depend on the initial state. 
The dissimilarity values show both high and low values, that correspond to have ``clusters'' of similar solutions.

On average, CNTF exhibits slightly better than \swotted, but the difference with \swotted (${\omega=1}$) is not significant across the dataset. 

For $\omega=3$ or $5$, we observe higher dissimilarity between the runs. Part of this increase can be explained by the metric used: cosine similarity tends to be higher for high-dimensional vectors (\ie larger phenotype sizes). This is because small differences in one dimension can lead to a significant decrease in cosine similarity, and the probability of such differences increases with the number of dimensions.

We conducted a qualitative analysis of the differences between the sets of phenotypes and found them to be almost the same. However,  we observed some discrepancies with a few extra or missing events. 
These events are recurrent in the data, but not necessarily related to a pathway. As the number of phenotypes is limited, it is better to include such events in a phenotype to improve reconstruction accuracy. Their weak association with other events can lead to variations between runs.

\begin{figure}[tb]
    \centering
    \includegraphics[width=\linewidth]{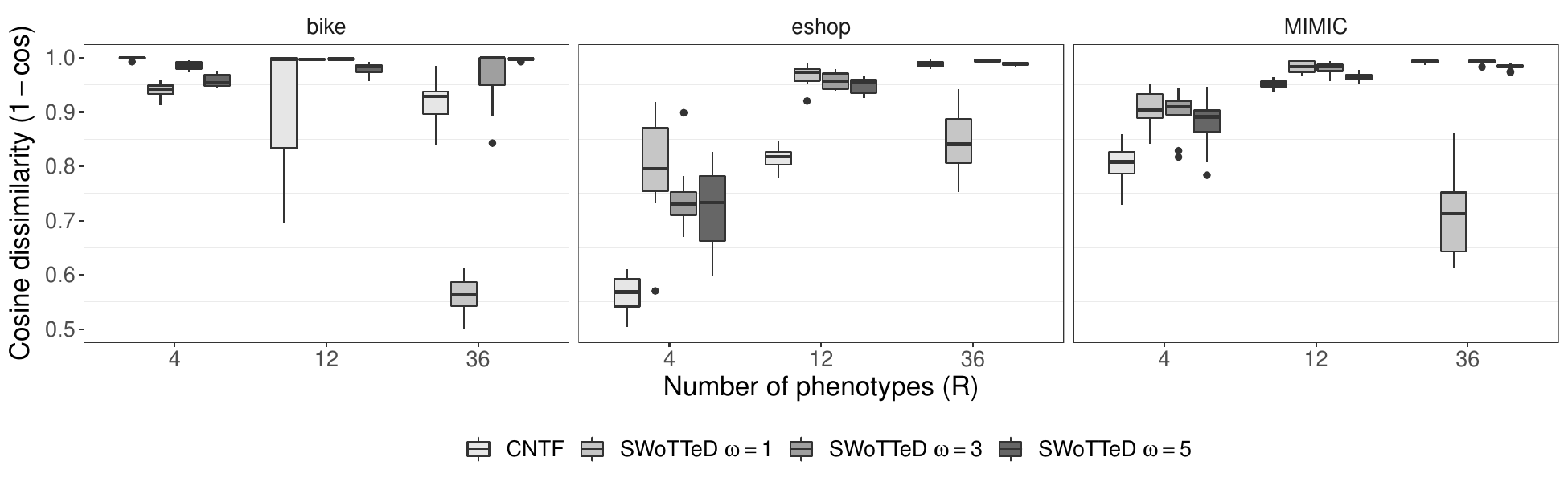}
    \caption{Cosine dissimilarity between pairs of phenotypes with respect to the phenotypes' size for \swotted, and CNTF. The higher the dissimilarity, the more diverse.}
    \label{fig:exp_diversity}
\end{figure}

\espace

Continuing our investigation of the similarities between phenotypes, we also evaluate the diversity of phenotypes within the sets of $R$ phenotypes. 
We computed the pairwise cosine dissimilarity between phenotypes within each extracted set by the different runs of \swotted. The objective is to evaluate each method's ability to extract a diverse set of phenotypes. 
It is worth noting that no orthogonality constraint, proposed by~\cite{kolda2001orthogonal} for instance,  is  directly implemented in \swotted (nor in CNTF). The diversity is expected as a side-effect of the reconstruction loss with a small set of phenotypes. For datasets with numerous latent behaviors, a diverse set of phenotypes ensures a better coverage of the data.

Figure~\ref{fig:exp_diversity} presents the distributions of cosine dissimilarity values. In this experiment, higher cosine dissimilarity values indicate greater diversity, which is desirable. 
We can notice the results are correlated to the analysis of uniqueness. Diverse sets of phenotypes corresponds to robust settings. This may be explained by the fact that the diverse sets extracted the complete set.

\espace

To summarize this section, we conclude that \swotted consistently converges toward sets of similar phenotypes on the real-world datasets for different runs. 
These sets contain diverse phenotypes, highlighting  \swotted's ability to discover non-redundant latent behaviors in temporal data. 
Despite these promising results, we recommend running \swotted multiple times on new datasets to enhance the confidence in the results.

\section{Case Study}\label{sec:casestudy} 

The previous experiments demonstrated that \swotted is able to accurately and robustly identify hidden phenotypes in synthetic data and to accurately reconstruct real-world data. 
In this section, we illustrate that the extracted temporal phenotypes are meaningful. 
For this purpose, we used \swotted on an EHR dataset from the Greater Paris University Hospitals and showed the outputted phenotypes to clinicians for interpretation.

The objective of this case study is to describe typical pathways of patients that have been admitted into intensive care units (ICU) during the first waves of COVID-19 in the Greater Paris region, France. The typical pathways are representative of treatment protocols that have actually been implemented.
Describing them may help hospitals to gain insight into their management of treatments during a crisis. 
In the context of COVID-19, we know that the most critical cases are patients with comorbidities (diabetes, hypertension, etc.). 
This complicates the analysis of these patients' care pathways because they blend multiple independent treatments. 
In such a situation, cutting edge tools for pathway analysis are helpful to disentangle the different treatments that have been delivered.

\begin{figure*}[t]
\centering
\includegraphics[width=0.19\textwidth]{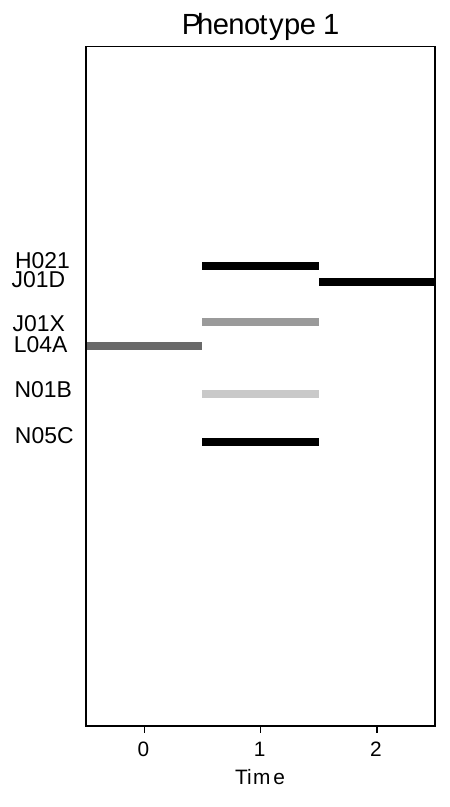}
\includegraphics[width=0.19\textwidth]{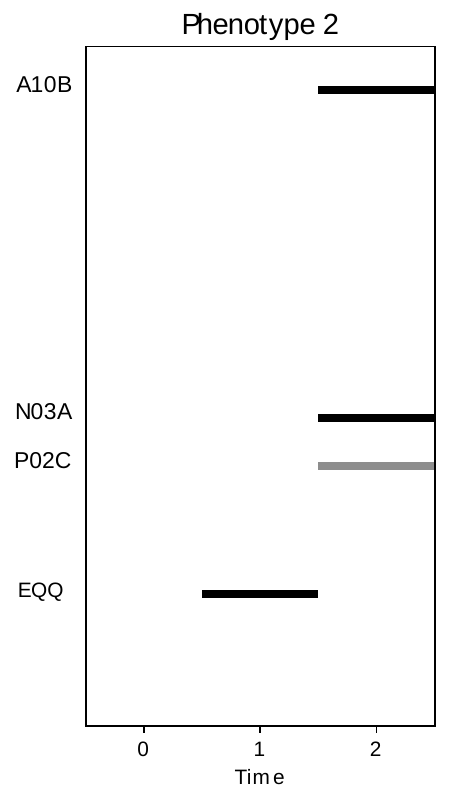}
\includegraphics[width=0.19\textwidth]{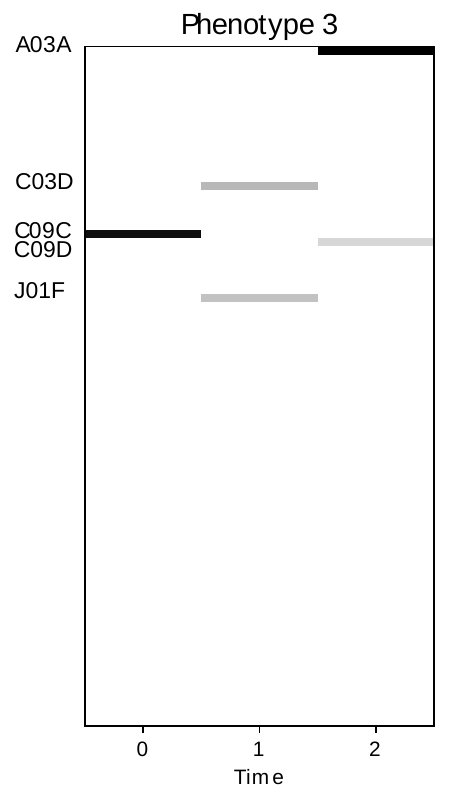} 
\includegraphics[width=0.19\textwidth]{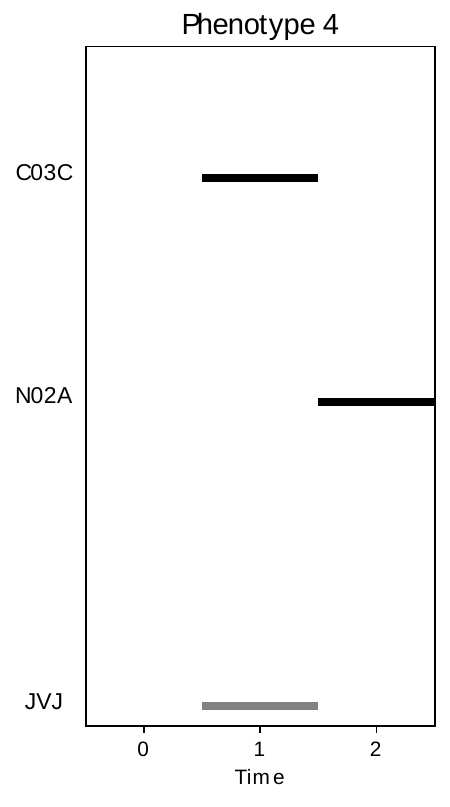} 
\includegraphics[width=0.19\textwidth]{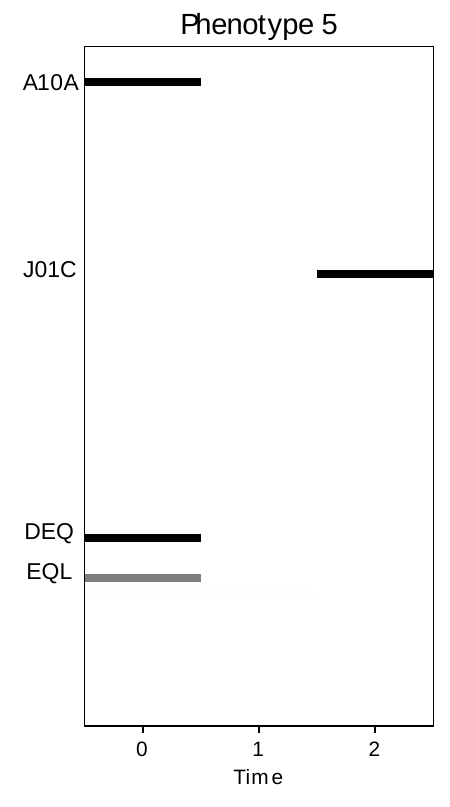}
\caption{Five phenotypes discovered for the 4th epidemic wave. Each gray cell represents the presence of a drug (in rows) at a relative time instant (in columns). 
The darker the cell, the higher the value. Cell values lie in the range $[0,1]$.} 
\label{fig:phw3:r8}
\end{figure*}

Care pathways of COVID-19 patients have been obtained from the data warehouse of the Greater Paris University Hospitals. 
We create one dataset per epidemic wave of COVID-19 for the first four waves. The periods of these waves are those officially defined by the French government. 
The patients selected for this study are adults (over 18 years old) with at least a positive PCR. 
For each patient, we create a binary matrix that represents the patient's care events (drugs deliveries and procedures) during the first $10$ days of his/her stay in the Intensive Care Unit (ICU). Epidemiologists selected $85$ types of care events ($58$ types of drugs and $27$ types of procedures) based on their frequency and relevance for COVID-19. 
Drugs are coded using the third level of ATC\footnote{ATC: Anatomical, Therapeutic and Chemical} codes and procedures are coded using the third level of CCAM\footnote{CCAM is the French classification of medical procedures} codes. 

In the following, we present the results obtained for the fourth wave (from 2021-07-05 to 2021-09-06) which holds $2,593$ patients and $21,325$ care events.\footnote{Details of the dataset preparation and all phenotypes for all the waves are available on Appendix A.4.3 and A.6.} We run \swotted to extract $R=8$ phenotypes of length $\omega=3$. We run $1,000$ epochs with a learning rate of $10^{-3}$.

Figure~\ref{fig:phw3:r8} illustrates five of the eight phenotypes extracted from the fourth wave.
At first glance, we can see that these phenotypes are sparse. 
This makes them almost easy to interpret: a phenotype is an arrangement of at most 7~care events, all with high weights. 
In addition, they make use of the time dimension: each phenotype describes the presence of care events on at least two different time instants. Thus, it demonstrates that the time dimension of a phenotype is meaningful in the decomposition. 
These phenotypes have been shown to a clinician for interpretation. 
It was confirmed that they reveal two relevant care combinations: some combinations of cares sketch the disease background of patients (hypertension, liver failure, etc.) while others are representative of treatment protocols. 
The phenotype 1 has been interpreted as a typical protocol for COVID-19. 
Indeed, L04A code referring to \textit{Tocilizumab} has become a standard drug to help patients with acute respiratory problems avoid mechanical ventilation. 
In this phenotype, clinicians detect a switch from the prophylactic delivery of \textit{Tocilizumab} (the first day) to a mechanical ventilation identified through the use of typical sedative drugs (N01B: \textit{Lidocaine}, J01X: \textit{Metronidazole} and N05C: \textit{Midazolam}). This switch, including the discontinuation of \textit{Tocilizumab} treatment, is a typical protocol. 
Nevertheless, further investigations are required to explain the presence of antibiotics (H02A: \textit{Prednisone} and J01D: \textit{Cefotaxime}).
Phenotype~5 illustrates a severe septic shock: a patient in this situation will be monitored (DEQ), injected with \textit{dopamine} (EQL) to induce cardiac activity, and with \textit{insulin} (A10A) to manage the patient's glycaemia. 
This protocol is commonly encountered in ICU, and is applied for COVID-19 patients in critical condition. 

The previously detailed phenotypes illustrate that \swotted disentangles generic ICU protocols and specific treatments for COVID-19. 
Other phenotypes have also been readily identified by clinicians as corresponding to treatments of patients having specific medical backgrounds. 
The details can be found in Appendix~\ref{sec:app:casestudy}. 
Their overall conclusion is that \swotted extracts relevant phenotypes that uncover some real practices.

\section{Conclusion and Perspectives}
The state-of-the-art tensor decomposition methods are limited to the extraction of phenotypes that only describe a combination of correlated features occurring the same day. 
In this article, we proposed a new tensor decomposition task that extracts temporal phenotypes, \ie, phenotypes that describe a temporal arrangement of events. 
We also propose \swotted, a tensor decomposition method dedicated to the extraction of temporal phenotypes.

\swotted has been intensively tested on synthetic and real-world datasets. 
The results show that it outperforms the current state-of-the-art tensor decomposition techniques on synthetic data by achieving the best reconstruction accuracies of both the input  data and the hidden phenotypes. 
The results on real-world data show that the reconstruction competes with state-of-the art methods, and extracts information through temporal phenotypes that is not captured by other approaches. 

In addition, we proposed a case study on COVID-19 patients to demonstrate the effectiveness of \swotted to extract meaningful phenotypes. This experiment illustrates the relevance of the temporal dimension to describe typical care protocols. 

The results of \swotted are very promising and open new research lines in machine learning, temporal phenotyping and care pathway analytics. For future work, we plan to extend \swotted to extract temporal phenotypes described over a variable window size. 
It would also be interesting to make the reconstruction more flexible for alternative applications. 
In particular, our temporal phenotypes are rigid sequential patterns: they describe the strict succession of days. This was expected to describe treatments in ICU, but some other applications might expect two consecutive days of a phenotype to match two days that are not strictly consecutive (with a gap in between). This is an interesting but challenging modification of the reconstruction which can be computationally expensive.
Finally, another possible improvement would be to use an AO-ADMM solver~\citep{huang2016flexible, roald2022admm}, which is known to increase the stability of tensor decomposition~\citep{Becker}.

\subsection*{Acknowledgements}
The authors thank the reviewers for their remarks and suggestions that greatly contributed to the improvement of this article. 
This work has been partly funded by the \href{https://www.bernoulli-lab.fr/en/project/chaire-ai-racles-us/}{AI-RACLES Chair}. 
The authors thank all contributors from the Clinical Data Warehouse of Greater Paris University Hospitals who work on making data available for research.
The case study on data from Greater Paris Hospital has an agreement from the Scientific and Ethical Committee of AP-HP (CSE-20-11 COVIPREDS, 27/03/2020).

\bibliographystyle{unsrtnat}
\bibliography{biblio.bib}

\begin{appendices}

\section{Details about Experiments}\label{sec:expesetup}

\subsection{Synthetic Dataset}
The code to generate synthetic data is available in the experiment repository: \url{https://gitlab.inria.fr/tguyet/swotted_experiments}. The file ``\texttt{gen\_data.py}'' implements the generation of synthetic data.  
We also provide a jupyter notebook ``\texttt{run\_SWoTTeD\_synthetic\_data.ipynb}'' that allows the reader to run readily \swotted on a synthetic dataset.

\subsection{MIMIC-IV Dataset}

\begin{figure}
\centering
\includegraphics[width=\textwidth]{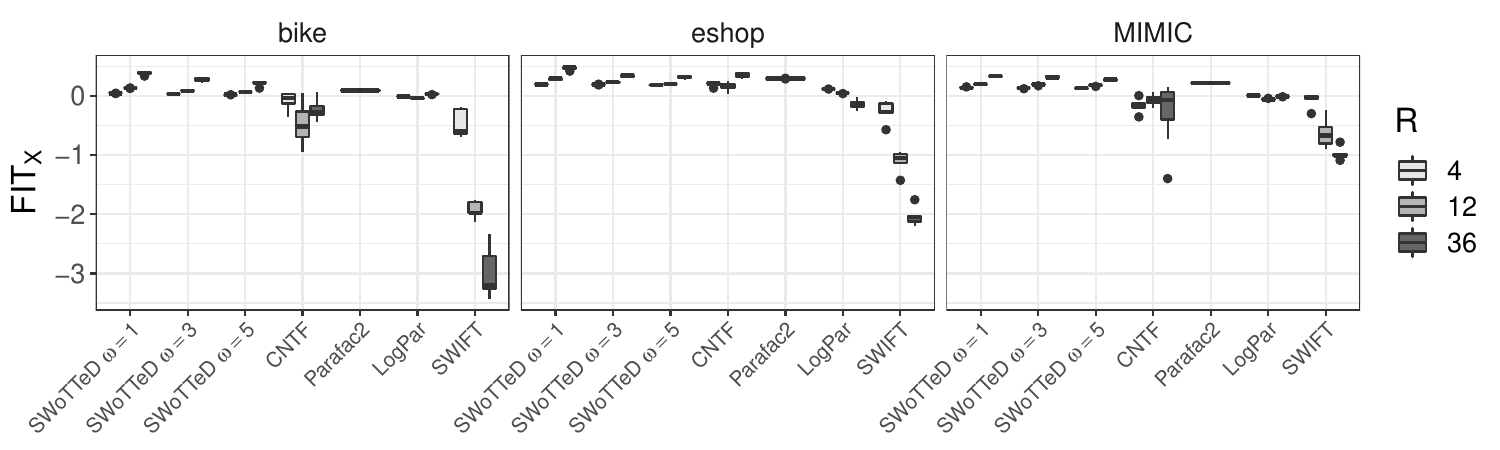}
\caption{Reconstruction error $FIT_X$ of \swotted ($\omega=1, 3, 5$) and its competitor on real-world datasets and for different values of $R$. }
\label{fig:result_RWD_FITX_full}
\end{figure}

The database can be accessed via this link \url{https://mimic.mit.edu/}. Once the database is set up in a Postgres database, the Jupyter notebook ``\texttt{CohortConstruction.ipynb}'' can be run to prepare the dataset: it builds the cohort of patients and the input tensors used in the experiments. 
We also provide a notebook ``\texttt{run\_SWoTTeD\_mimic.ipynb}'' to guide the reader to execute \swotted on MIMIC-IV.

The availability of this data preparation enables anyone to create the exact dataset we used for our experiments and  to compare his/her own results with ours.

\section{Additional Results}

\subsection{Impact of Normalization}\label{sec:annex:normalization}
\begin{figure}[tb]
\centering
\includegraphics[width=\textwidth]{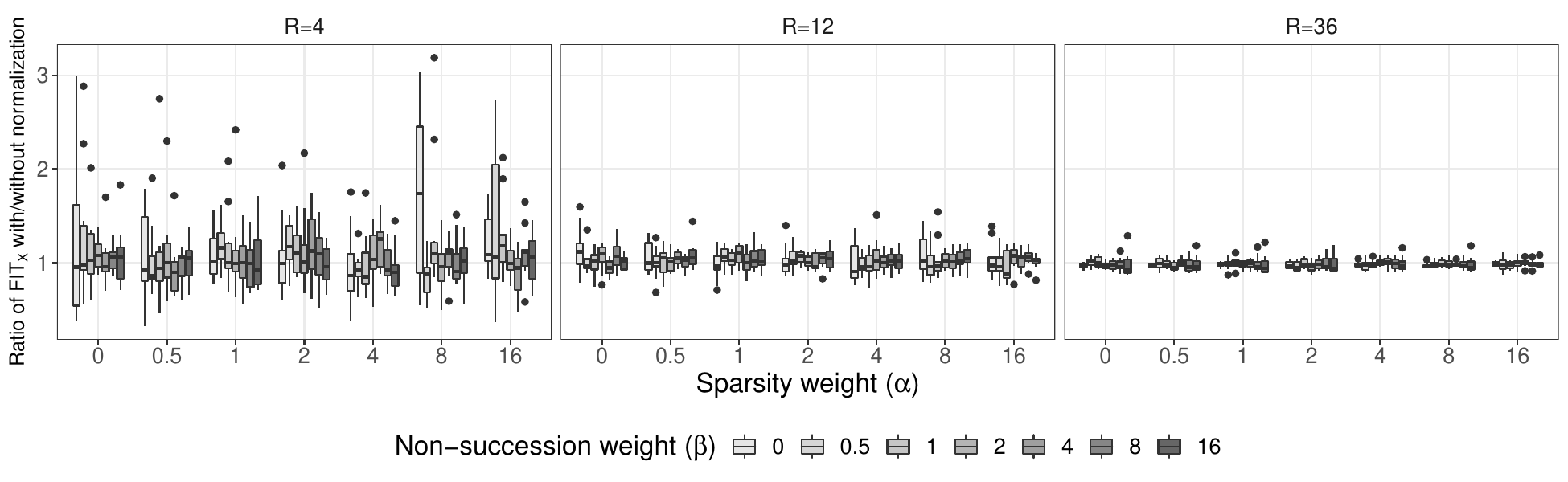}
\caption{Ratio of the $FIT_X$ metric with and without normalization of \swotted on synthetic datasets. 
A ratio above $1$ indicates a benefit of using normalization on the reconstruction accuracy.
}
\label{fig:result_synth_normalization}
\end{figure}

Similar experiments were conducted without normalization. 
The Figure~\ref{fig:result_synth_normalization} depicts the ratio of~$FIT_X$ with and without normalization.
We observe that the values are very close to 1. On average, normalization appears to increase the $FIT$~measure (ratio above 1), but it is not significant. 
We conclude from this experiment that normalization does not clearly improves the reconstruction accuracy neither the quality of the extracted phenotypes. 
Therefore, we recommend keeping the normalization constraint because it guarantees that $\mathcal{W}$'s values are limited to the range~$[0,1]$ and they can be interpreted as probabilities. 

\subsection{Comparison of FIT Computed on Train or Test Datasets}

The objective of this experiment is to show that the phenotypes discovered does not overfit the train dataset. 
To demonstrate this, we computed the $FIT_X$ values on test and train sets. 

\begin{figure}[tb]
\centering
\includegraphics[width=\textwidth]{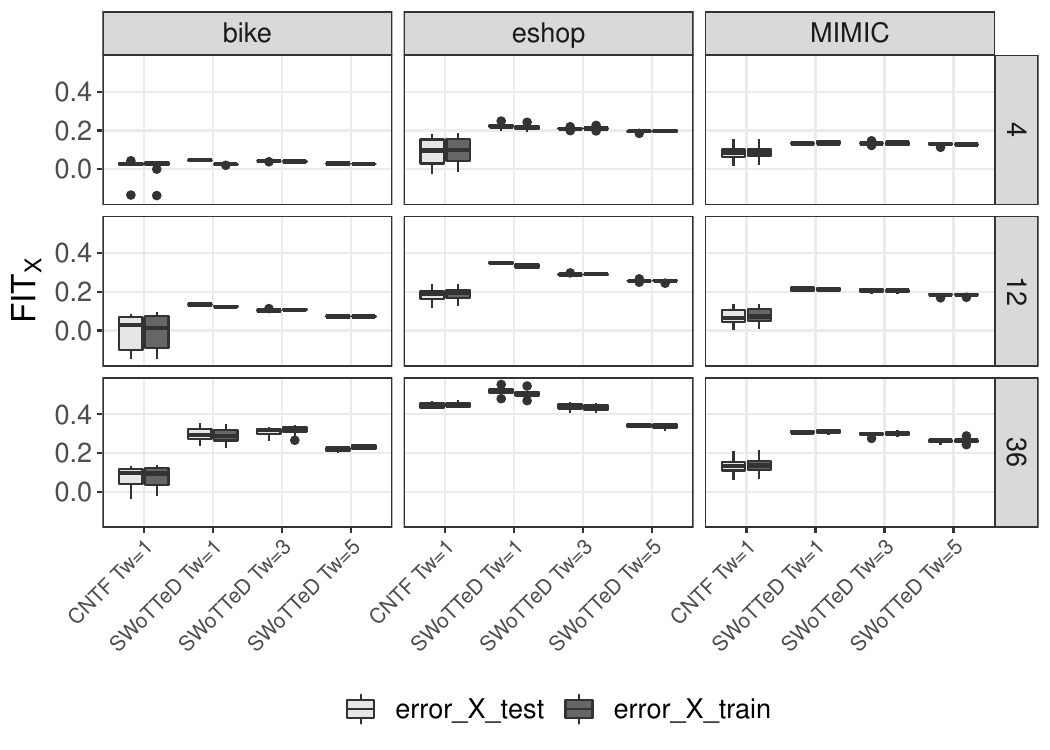}
\caption{$FIT_X$ metric computed on test and train datasets for realworld datasets.
}
\label{fig:result_rwd_traintest}
\end{figure}

Figure~\ref{fig:result_rwd_traintest} presents the results obtained for our four real-world datasets. 
We observe that the $FIT_X$ computed on train and test datasets are on average similar both for \swotted and CNTF.
It shows that neither \swotted nor CNTF overfits the train dataset. This result contributes to show that the extracted phenotypes are meaningful.

\subsection{Comparison of Time Efficiency of \swotted vs {\sc FastSWoTTeD}}\label{sec:appendix:cmp_fastswotted}

We remind that {\sc FastSWoTTeD} is an implementation of \swotted for datasets having all sequences with the same length. In this case, it is possible to increase the computational efficiency of the implementation with vectorization techniques. We also remind that we did not use GPUs to accelerate the training.

In this experiment we generated 200 synthetic sequences with all same lengths ($T=20$) and $40$ features. We run the two models on the same datasets and we measure the training time, and the $FIT_X$ on a test set.

\begin{figure}[tb]
\centering
\includegraphics[width=.7\textwidth]{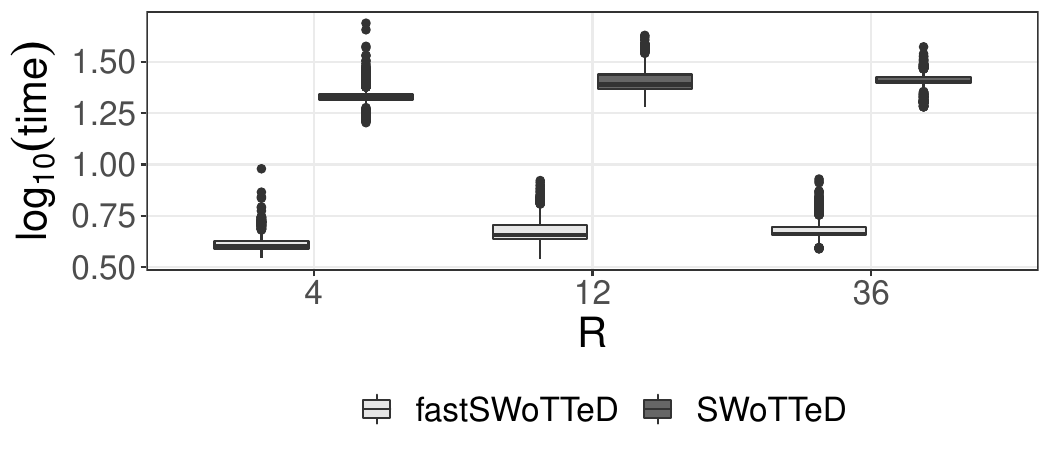}
\caption{$\log_{10}(time)$ (in seconds) for training \swotted and {\sc FastSWoTTeD} models.}
\label{fig:result_time_fastswotted}
\end{figure}

Figure~\ref{fig:result_time_fastswotted} depicts the computing times with respect to the number of hidden phenotypes ($R$). The boxplot averages the result for different model settings. We clearly see that {\sc FastSWoTTeD} is more than one order of magnitude faster than \swotted. 

\begin{figure}[tb]
\centering
\includegraphics[width=.7\textwidth]{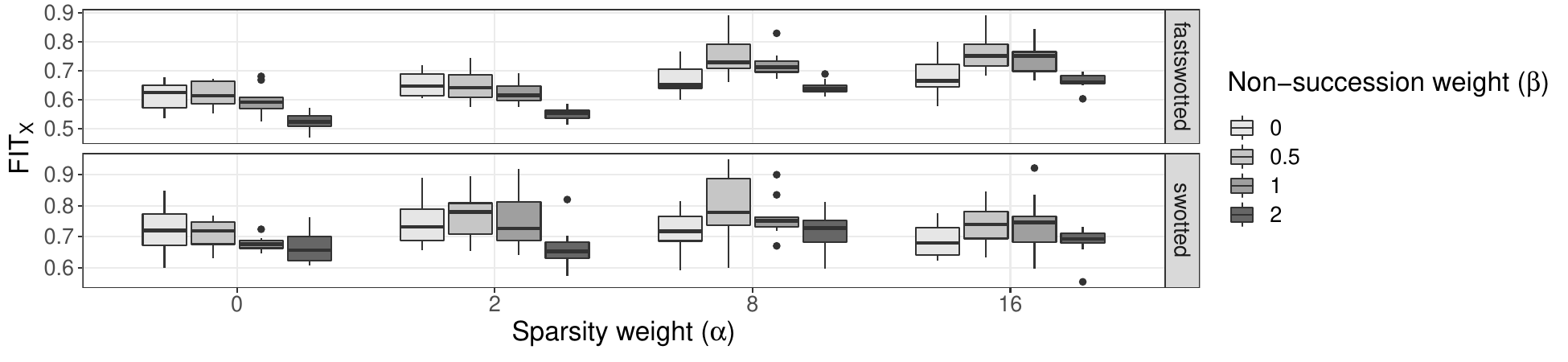}
\caption{$FIT_X$ on synthetic datasets obtained with \swotted and {\sc FastSWoTTeD} models.}
\label{fig:result_fit_fastswotted}
\end{figure}

Figure~\ref{fig:result_fit_fastswotted} depicts the $FIT_X$ obtained with the two implementations of \swotted. We observe that the two implements have achieves the same $FIT_X$. The difference can be explained by the randomness in the training process, combined with small floating point errors that propagate differences between the decompositions. Statistically, the differences are not significant between these two models.

\section{Computational Complexity of the Reconstruction Loss}\label{sec:app:complexity}

Let $\mathcal{X}$ be a dataset with $K$ patients, $n$ medical features, and a maximum hospital stay length of $\bar{T}$; let $R$ be the desired number of phenotypes described over a temporal window of size $\omega$. The complexity to compute the reconstruction $\mathcal{P} \circledast \mathbf{W}^{(k)}$ is given by:
$$O\left(K\times\bar{T}\times R\times n\times \omega\right).$$
It follows that the complexity to compute the error $\mathcal{L}^{\mathsmaller \circledast}$ (see Equation~\ref{eq:SWloss}) is as follows:
$$O\left(\left(R\times \omega + 1\right)\times K\times\bar{T}\times n\right).$$

The complexity of the loss of \swotted (see Equation~\ref{eq:SW_loss}) includes the complexity of the regularization terms, as outlined below:
\begin{itemize}
\item The complexity of the $L_1$ norm is $O\left(n \times \omega \times R\right)$ (i.e., the size of $\mathcal{P}$).
\item The complexity of the non-succession regularization term $\mathcal{S}(\mathbf{W}^{(k)})$ is $O\left(K \times R \times \bar{T} \times \omega\right)$.
\end{itemize}

In comparison to non-temporal tensor decomposition techniques (e.g., CNTF), the complexity of \swotted considers the $\omega$~factor that corresponds to the temporal width of phenotypes. While \swotted is more expressive, it requires more computing resources than CNTF\footnote{Here, we consider CNTF model without temporal regularization. Indeed, temporal regularization of CNTF is based on an internal LSTM architecture that is time consuming to train.}. On the other hand, the reconstruction and non-succession term are based on convolution operators that benefit from material optimization. Thus, we anticipate that the efficiency of the optimization process will be preserved.

\vspace{10pt}

\swotted is based on an optimization procedure for training the model. Each epoch comprises the computation of the loss $\ell$, and its derivatives. Considering that \swotted is based on automatic differentiation, the time complexity to evaluate the derivative is proportional to the computation of the loss~\citep{1fc237b2-9136-3c8d-b48d-5eaa15ea74f6}.

\section{Case Study Details}\label{sec:app:casestudy}

\subsection{Dataset Preparation}
In our case study, we used data from the Greater Paris University Hospitals.
For the sake of health data security and privacy, these data have been pseudo-anonymized and it is not possible to export them out of the hospital infrastructure. Then, all experiments on the AP-HP dataset were conducted on hospital servers through secured access.

In addition, all the experiments have been approved by the Ethical committee of AP-HP which is in charge to inform patients about their data usage for research purposes, to obtain their informed consent and to validate the appropriateness of data access by researchers. 

In accordance with epidemiologists, we created one dataset per epidemic wave of COVID-19 for the first four waves. The periods of these waves are those officially defined by the French government:\footnote{\url{https://www.insee.fr/fr/statistiques/5432509?sommaire=5435421}} 
\begin{itemize}
\item wave 1: \textit{2020-03-01} - \textit{2020-07-06}, $8,561$ patients, $81,517$ care events
\item wave 2: \textit{2020-07-07} - \textit{2021-01-04}, $10,444$ patients, $91,824$ care events
\item wave 3: \textit{2021-01-05} - \textit{2021-07-05}, $14,667$ patients, $139,045$ care events
\item wave 4: \textit{2021-07-05} - \textit{2021-09-06}, $2,593$ patients, $21,325$ care events
\end{itemize}

The patients selected for this study are:
\begin{itemize}
\item adults (over 18 years old) 
\item at least a positive PCR\footnote{Polymerase Chain Reaction} test to witness the COVID-19 status of the patient
\item admitted in an Intensive Care Unit (ICU). ICU includes here the historical and \textit{de novo} units. The latter have been by set up to respond urgently to the massive arrival of patients. 
\end{itemize} 

For each patient we create a binary matrix that represents its care pathway.

We collected medical procedures and drugs delivered during the 10 first days in an Intensive Care Unit (ICU). All care pathways start from the entrance in an ICU. This means that all cares received at hospital before the entrance in ICU are ignored.  If a patient is admitted several times in an ICU, each visit is independent of the others. 

All care pathways of length strictly larger than 10 are truncated. Indeed, no further information is expected beyond the tenth day. In the final dataset, 60\% of the care pathways are truncated. In addition, stays shorter than 2 days are discarded.

Drugs are coded using ATC \footnote{ATC: Anatomical, Therapeutical and Chemical} codes and procedures are coded using CCAM\footnote{CCAM is the French classification of medical procedures} codes. 
CCAM is the French Common Classification of medical procedures. Each code is a type of medical event in the input tensor. 
The date of occurrence of an event is determined by the number of days before the beginning of mechanical ventilation. 
More specifically, it means that the sampling rate is one day. 
In the data warehouse, procedure and drug deliveries are timestamped but, according to clinicians, these timestamps are not reliable (for instance, care deliveries are entered in the system at once at the end of the day). 
If the length of stay is lower than 10 days (whatever the discharge reason), then the temporal dimension of its matrix is its length of stay. More specifically, all matrices do not have the same temporal dimensions.

We first selected the 500 most frequent drugs and the 200 most frequent procedures over a very large space. Then, drugs have been clustered at the 3rd ATC level, and procedures at the 3th level of CCAM coding hierarchy. Indeed, different drug names may correspond to the same kind of cares. Thus, we group drugs using the coding hierarchy of the ATC and of the CCAM. 
A second manual selection has been done by physicians to keep only the most relevant features in the context of COVID-19. 
In the end, $85$ types of care events ($58$ types of drugs and $27$ types of procedures) have been selected by physicians based on their frequency and relevance for COVID-19. 

It is worth noticing that we do not have access to a reliable information about the duration of a drug prescription. For instance, we noticed that antibiotic drugs are delivered to patients on a single day each time, but not several days in a row. Nonetheless, clinicians indicates that antibiotic are always delivered for several days to be effective. 
This is a limitation to identify temporal phenotypes, but we prefer to not pre-process the data to prevent from having too questionable results.

The following table presents the parameters used for the results presented in Section~\ref{sec:results:aphp}.

\begin{center}
\begin{tabular}{lc}
\toprule
Parameter & Value \\
\midrule
$R$ & $8$ or $20$\\
$\omega$ & $3$ or $5$\\
$\alpha$ & $0.5$\\
$\beta$ & $0.5$\\[2pt]\hdashline 
$\lambda$ & $10^{-3}$ \\
epochs & $1,000$\\
batch size & $64$\\
\bottomrule
\end{tabular}
\end{center}

\subsection{Additional Results on Case Study}\label{sec:results:aphp}
In this section, we present the results obtained for the four first waves of the COVID-19 pandemic. We start by providing results about the reconstruction accuracy for several settings of \swotted.  
Then, we illustrate the reconstruction of some patient matrices. 
Finally, we present the graphical and detailed descriptions of the phenotypes obtained with $R=8$ phenotypes and $\omega=3$. 

Note that these results need additional investigations by clinicians to provide reliable interpretations. For this reason, we do not provide more detailed interpretations of phenotypes in this document.

\subsubsection{Reconstruction Accuracy}

Table~\ref{tab:aphp:fit} gives the $FIT_X$ values obtained on the four AP-HP datasets (four first waves of the pandemic). 
The values are close to the ones obtained on the MIMIC-IV dataset. These values are positive, indicating that the reconstruction is reasonably good on average. 

\begin{table}[h]
\caption{Reconstruction accuracy measured by $FIT_X$ of the AP-HP datasets (one per wave) with different settings of \swotted. Bold values are the bests.}
\label{tab:aphp:fit}
\centering
\begin{tabular}{lccc}
\toprule
Wave & $R=8$ &$R=20$ &$R=20$ \\
 & $\omega=3$ &$\omega=3$ &$\omega=5$ \\
\midrule
w1 & 0.17& \textbf{0.27} & 0.22\\
w2 & 0.15& \textbf{0.25}& 0.23\\
w3 & 0.15& \textbf{0.23}& 0.23 \\
w4 & 0.15& 0.25& \textbf{0.27} \\
\bottomrule
\end{tabular}
\end{table}

We can notice that \swotted achieves better reconstruction with $R=20$ than with $R=8$. Indeed, with more phenotypes, the decomposition becomes easier. 
The drawback of having more phenotypes is to discover simpler patterns, \ie involving fewer events. 

Finally, increasing the size of the temporal phenotypes does not improve the reconstruction. Larger temporal phenotypes means a more constrained reconstruction.

\subsubsection{Reconstructions}
\begin{figure*}
\includegraphics[width=.48\linewidth]{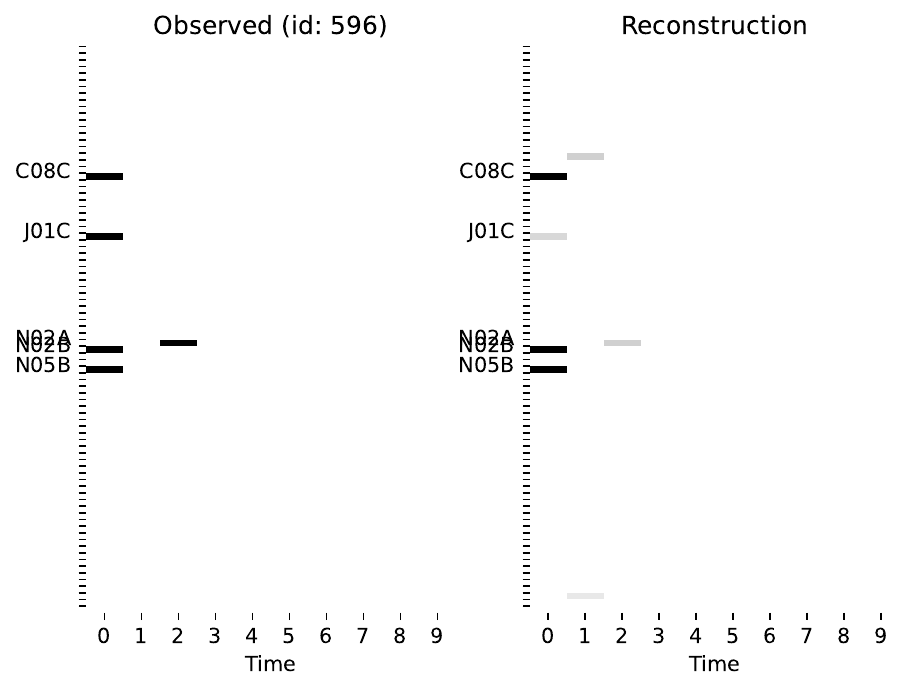}\hfill
\includegraphics[width=.48\linewidth]{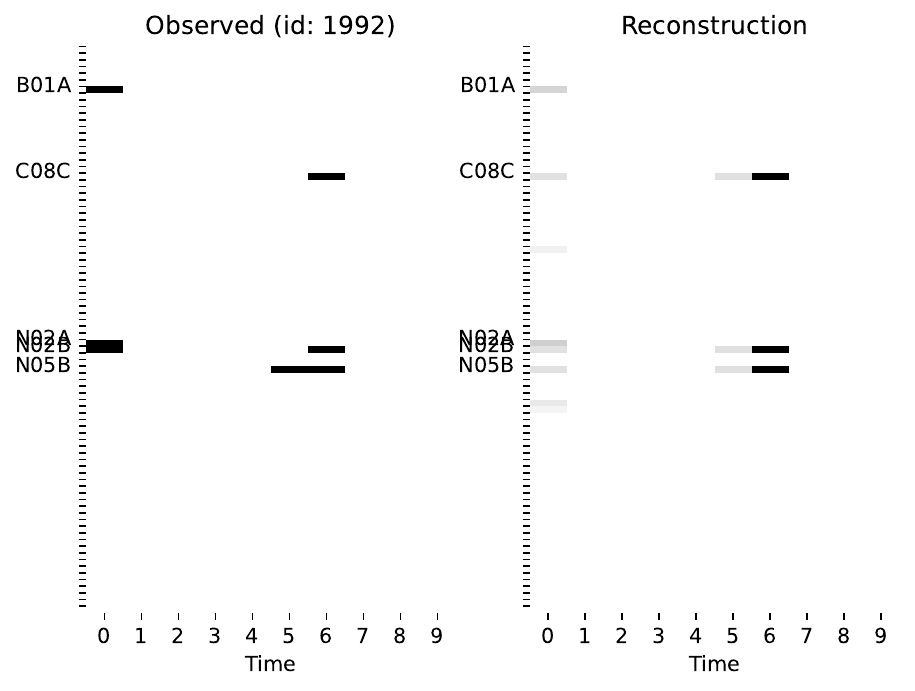}
\caption{Examples of patient matrix reconstruction (with $R=8$ and $\omega=3$). For each patient, the original matrix is on the left, the reconstructed matrix is on the right.}
\label{fig:aphp:reconstruction}
\end{figure*}

Figure~\ref{fig:aphp:reconstruction} illustrates the reconstructions of two patient's matrices. The patients are in the dataset of the fourth wave, and the reconstruction use $R=8$ and $\omega=3$. 

These reconstructions illustrate the behaviour we observed: 1) a reconstruction contains all the events of the input matrix, 2) some additional events appear in the reconstruction.
For the patient on the left ($id:596$), the reconstruction contains, with significant weights, the five care events of the initial matrix. But, an additional event appears in the reconstruction (with a low weight). 
For the patient on the right ($id:1992$), the first day of the reconstruction contains a lot of lightly weighted events, but it reconstructs well the complex combination at time $6$.

A possible explanation of this behaviour can be found in the weak quality of the data. As we notice, there may miss some events. The additional events in the reconstruction may witness such misinformed care deliveries. This hypothesis remains to be investigated.

\subsubsection{Phenotypes of the First Wave}

\begin{landscape}
\begin{figure*}[ph]
\centering
\includegraphics[width=\linewidth]{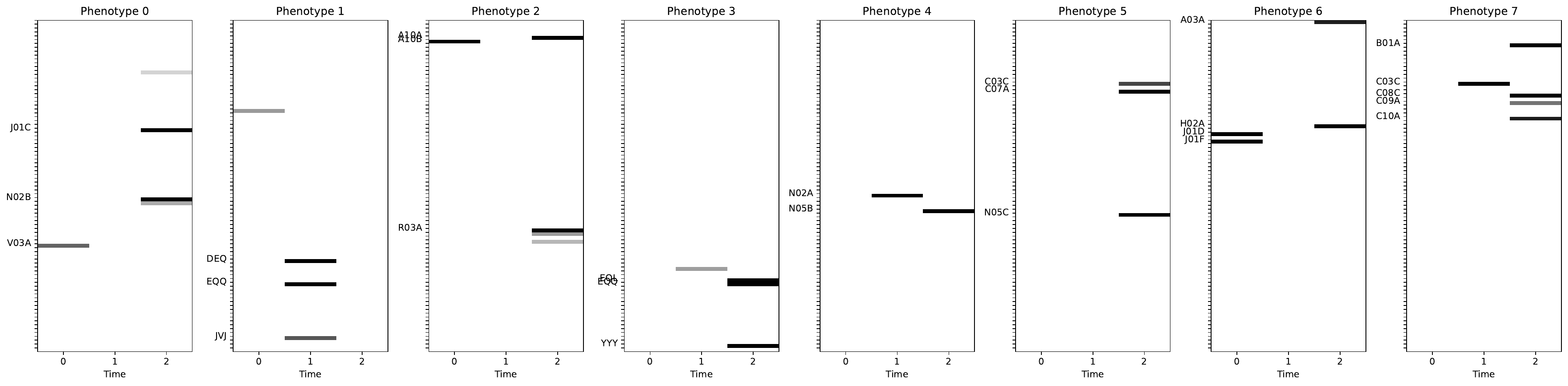}
\caption{\scriptsize{Graphical representation of the phenotypes of the first wave.}} 
\label{fig:phenotypes:wave1}

\includegraphics[width=\linewidth]{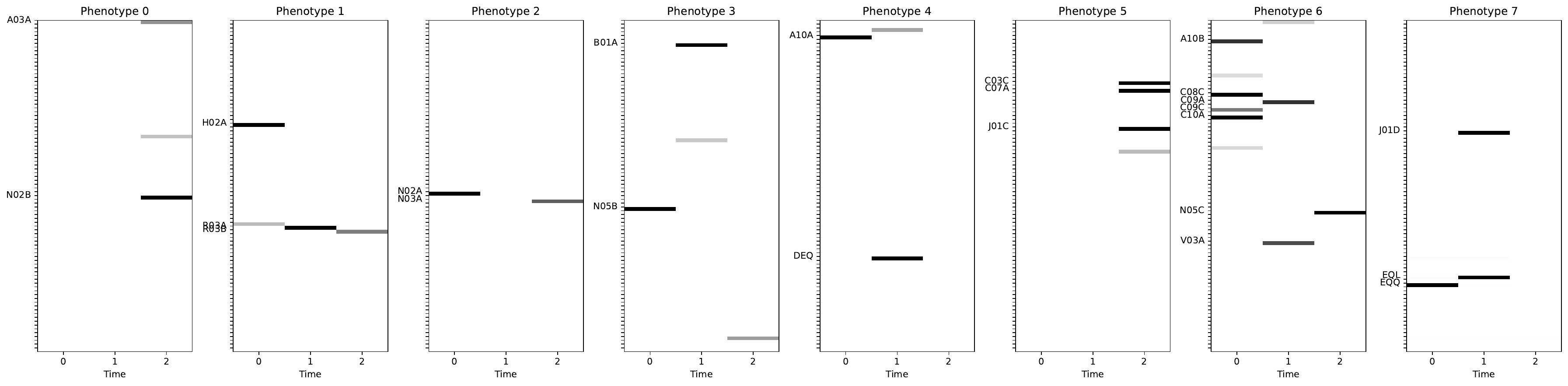}
\caption{Graphical representation of the phenotypes of the second wave.}
\label{fig:phenotypes:wave2}

\end{figure*}
\end{landscape}

Figure~\ref{fig:phenotypes:wave1} illustrates the phenotypes that have been extracted. These phenotypes are detailed in Table~\ref{tab:appendix:firstwave}.

\begin{table}[p]
\scriptsize \begin{tabular}{lp{6cm}c:c:c}
\multicolumn{4}{l}{Phenotype 1}\\
\toprule
Event code & Description &  \multicolumn{3}{l}{Time}\\
\midrule\textsf{C02C} & \textit{\small Urapidil, hypertension}& 0.00& 0.00& 0.17\\
\textsf{J01C} & \textit{\small Amoxicilline}& 0.00& 0.00& 1.00\\
\textsf{N02B} & \textit{\small Paracetamol}& 0.00& 0.00& 1.00\\
\textsf{N03A} & \textit{\small Pregabaline}& 0.00& 0.00& 0.33\\
\textsf{V03A} & \textit{\small Oxygen}& 0.61& 0.00& 0.00\\
\bottomrule
\end{tabular}

\scriptsize \begin{tabular}{lp{6cm}c:c:c}
\multicolumn{4}{l}{Phenotype 2}\\
\toprule
Event code & Description &  \multicolumn{3}{c}{Days}\\
\midrule\textsf{C09C} & \textit{\small Irbesartan, ARB}& 0.39& 0.00& 0.00\\
\textsf{DEQ} & \textit{\small Electrocardiogram}& 0.00& 1.00& 0.00\\
\textsf{EQQ} & \textit{\small Cardiac monitoring}& 0.00& 1.00& 0.00\\
\textsf{JVJ} & \textit{\small Dialysis}& 0.00& 0.66& 0.00\\
\bottomrule
\end{tabular}

\noindent
\scriptsize \begin{tabular}{lp{6cm}c:c:c}
\multicolumn{4}{l}{Phenotype 3}\\
\toprule
Event code & Description &  \multicolumn{3}{c}{Days}\\
\midrule\textsf{A10A} & \textit{\small Insuline}& 0.00& 0.00& 1.00\\
\textsf{A10B} & \textit{\small Metformine}& 1.00& 0.00& 0.00\\
\textsf{R03A} & \textit{\small Adrenergics, inhalants}& 0.00& 0.00& 1.00\\
\textsf{R03B} & \textit{\small Drugs for obstructive airway diseases}& 0.00& 0.00& 0.39\\
\textsf{R06A} & \textit{\small Antihistamines}& 0.00& 0.00& 0.28\\
\bottomrule
\end{tabular}

\noindent
\scriptsize \begin{tabular}{lp{6cm}c:c:c}
\multicolumn{4}{l}{Phenotype 4}\\
\toprule
Event code & Description &  \multicolumn{3}{c}{Days}\\
\midrule\textsf{ENL} & \textit{\small Monitoring of intra-arterial pressure}& 0.00& 0.38& 0.00\\
\textsf{EQL} & \textit{\small Dopamine}& 0.00& 0.00& 1.00\\
\textsf{EQQ} & \textit{\small Cardiac monitoring}& 0.00& 0.00& 1.00\\
\textsf{YYY} & \textit{\small Rescucitation procedures}& 0.00& 0.00& 1.00\\
\bottomrule
\end{tabular}

\noindent
\scriptsize \begin{tabular}{lp{6cm}c:c:c}
\multicolumn{4}{l}{Phenotype 5}\\
\toprule
Event code & Description &  \multicolumn{3}{c}{Days}\\
\midrule\textsf{N02A} & \textit{\small Morphines}& 0.00& 1.00& 0.00\\
\textsf{N05B} & \textit{\small Hydroxizine, sedation}& 0.00& 0.00& 1.00\\
\bottomrule
\end{tabular}

\noindent
\scriptsize \begin{tabular}{lp{6cm}c:c:c}
\multicolumn{4}{l}{Phenotype 6}\\
\toprule
Event code & Description &  \multicolumn{3}{c}{Days}\\
\midrule\textsf{C03C} & \textit{\small Furosemide}& 0.00& 0.00& 0.73\\
\textsf{C07A} & \textit{\small Bisoprolol}& 0.00& 0.00& 1.00\\
\textsf{N05C} & \textit{\small Midazolam, sedation}& 0.00& 0.00& 1.00\\
\bottomrule
\end{tabular}

\noindent
\scriptsize \begin{tabular}{lp{6cm}c:c:c}
\multicolumn{4}{l}{Phenotype 7}\\
\toprule
Event code & Description &  \multicolumn{3}{c}{Days}\\
\midrule\textsf{A03A} & \textit{\small Phloroglucinol}& 0.00& 0.00& 0.88\\
\textsf{H02A} & \textit{\small Prednisone, antibiotic}& 0.00& 0.00& 1.00\\
\textsf{J01D} & \textit{\small Cefotaxime}& 1.00& 0.00& 0.00\\
\textsf{J01F} & \textit{\small Azithromicine}& 1.00& 0.00& 0.00\\
\bottomrule
\end{tabular}

\noindent
\scriptsize \begin{tabular}{lp{6cm}c:c:c}
\multicolumn{4}{l}{Phenotype 8}\\
\toprule
Event code & Description &  \multicolumn{3}{c}{Days}\\
\midrule\textsf{B01A} & \textit{\small Antithrombotic agents}& 0.00& 0.00& 1.00\\
\textsf{C03C} & \textit{\small Furosemide}& 0.00& 1.00& 0.00\\
\textsf{C08C} & \textit{\small Amlodipine}& 0.00& 0.00& 1.00\\
\textsf{C09A} & \textit{\small Ramipril}& 0.00& 0.00& 0.54\\
\textsf{C10A} & \textit{\small Statine}& 0.00& 0.00& 0.89\\
\bottomrule
\end{tabular}
\caption{Detailed description of phenotypes of the first wave.}
\label{tab:appendix:firstwave}
\end{table}

\subsubsection{Phenotypes of the Second Wave}
Figure~\ref{fig:phenotypes:wave2} illustrates the phenotypes that have been extracted. These phenotypes are detailed in Table~\ref{tab:appendix:secondwave}.

\begin{table}[p]
\scriptsize \begin{tabular}{lp{6cm}c:c:c}
\multicolumn{4}{l}{Phenotype 1}\\
\toprule
Event code & Description &  \multicolumn{3}{c}{Days}\\
\midrule\textsf{A03A} & \textit{\small Phloroglucinol}& 0.00& 0.00& 0.42\\
\textsf{J01E} & \textit{\small Sulfonamides and trimethoprim}& 0.00& 0.00& 0.24\\
\textsf{N02B} & \textit{\small Paracetamol}& 0.00& 0.00& 1.00\\
\bottomrule
\end{tabular}

\scriptsize \begin{tabular}{lp{6cm}c:c:c}
\multicolumn{4}{l}{Phenotype 2}\\
\toprule
Event code & Description &  \multicolumn{3}{c}{Days}\\
\midrule\textsf{H02A} & \textit{\small Prednisone, antibiotic}& 1.00& 0.00& 0.00\\
\textsf{P02C} & \textit{\small Ivermectine}& 0.26& 0.00& 0.00\\
\textsf{R03A} & \textit{\small Adrenergics, inhalants}& 0.00& 1.00& 0.00\\
\textsf{R03B} & \textit{\small Drugs for obstructive airway diseases}& 0.00& 0.00& 0.50\\
\bottomrule
\end{tabular}

\scriptsize \begin{tabular}{lp{6cm}c:c:c}
\multicolumn{4}{l}{Phenotype 3}\\
\toprule
Event code & Description &  \multicolumn{3}{c}{Days}\\
\midrule\textsf{N02A} & \textit{\small Morphines}& 1.00& 0.00& 0.00\\
\textsf{N03A} & \textit{\small Pregabaline}& 0.00& 0.00& 0.63\\
\bottomrule
\end{tabular}

\scriptsize \begin{tabular}{lp{6cm}c:c:c}
\multicolumn{4}{l}{Phenotype 4}\\
\toprule
Event code & Description &  \multicolumn{3}{c}{Days}\\
\midrule\textsf{B01A} & \textit{\small Antithrombotic agents}& 0.00& 1.00& 0.00\\
\textsf{J01F} & \textit{\small Azithromicine}& 0.00& 0.22& 0.00\\
\textsf{N05B} & \textit{\small Hydroxizine, sedation}& 1.00& 0.00& 0.00\\
\textsf{JVJ} & \textit{\small Dialysis}& 0.00& 0.00& 0.38\\
\bottomrule
\end{tabular}

\scriptsize \begin{tabular}{lp{6cm}c:c:c}
\multicolumn{4}{l}{Phenotype 5}\\
\toprule
Event code & Description &  \multicolumn{3}{c}{Days}\\
\midrule\textsf{A06A} & \textit{\small Constipation}& 0.00& 0.35& 0.00\\
\textsf{A10A} & \textit{\small Insuline}& 1.00& 0.00& 0.00\\
\textsf{DEQ} & \textit{\small Electrocardiogram}& 0.00& 1.00& 0.00\\
\bottomrule
\end{tabular}

\scriptsize \begin{tabular}{lp{6cm}c:c:c}
\multicolumn{4}{l}{Phenotype 6}\\
\toprule
Event code & Description &  \multicolumn{3}{c}{Days}\\
\midrule\textsf{C03C} & \textit{\small Furosemide}& 0.00& 0.00& 1.00\\
\textsf{C07A} & \textit{\small Bisoprolol}& 0.00& 0.00& 1.00\\
\textsf{J01C} & \textit{\small Amoxicilline}& 0.00& 0.00& 1.00\\
\textsf{J01X} & \textit{\small Metronidazole}& 0.00& 0.00& 0.27\\
\bottomrule
\end{tabular}

\scriptsize \begin{tabular}{lp{6cm}c:c:c}
\multicolumn{4}{l}{Phenotype 7}\\
\toprule
Event code & Description &  \multicolumn{3}{c}{Days}\\
\midrule\textsf{A03A} & \textit{\small Phloroglucinol}& 0.00& 0.19& 0.00\\
\textsf{A10B} & \textit{\small Metformine}& 0.80& 0.00& 0.00\\
\textsf{C03A} & \textit{\small Hydrochlorothiazide}& 0.14& 0.00& 0.00\\
\textsf{C08C} & \textit{\small Amlodipine}& 1.00& 0.00& 0.00\\
\textsf{C09A} & \textit{\small Ramipril}& 0.00& 0.80& 0.00\\
\textsf{C09C} & \textit{\small Irbesartan, ARB}& 0.52& 0.00& 0.00\\
\textsf{C10A} & \textit{\small Statine}& 1.00& 0.00& 0.00\\
\textsf{J01M} & \textit{\small Ciprofloxacine}& 0.16& 0.00& 0.00\\
\textsf{N05C} & \textit{\small Midazolam, sedation}& 0.00& 0.00& 1.00\\
\textsf{V03A} & \textit{\small Oxygen}& 0.00& 0.70& 0.00\\
\bottomrule
\end{tabular}

\scriptsize \begin{tabular}{lp{6cm}c:c:c}
\multicolumn{4}{l}{Phenotype 8}\\
\toprule
Event code & Description &  \multicolumn{3}{c}{Days}\\
\midrule\textsf{J01D} & \textit{\small Cefotaxime}& 0.00& 1.00& 0.00\\
\textsf{DEQ} & \textit{\small Electrocardiogram}& 0.00& 0.01& 0.00\\
\textsf{EQL} & \textit{\small Dopamine}& 0.00& 1.00& 0.00\\
\textsf{EQQ} & \textit{\small Cardiac monitoring}& 1.00& 0.00& 0.00\\
\bottomrule
\end{tabular}
\caption{Detailed description of phenotypes of the second wave.}
\label{tab:appendix:secondwave}
\end{table}


\subsubsection{Phenotypes of the Third Wave}
\begin{landscape}
\begin{figure*}[ph]
\centering
\includegraphics[width=\linewidth]{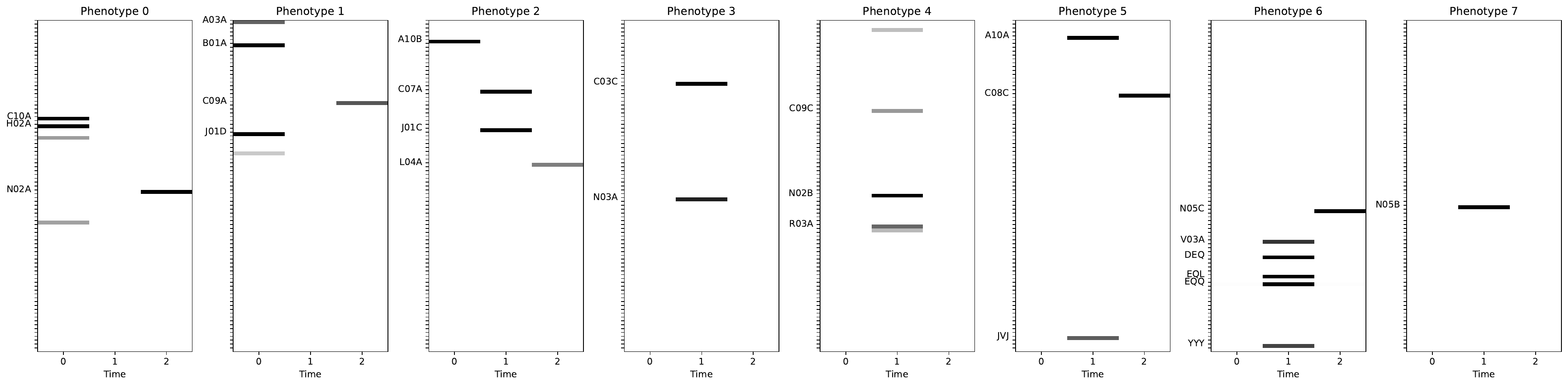}
\caption{Graphical representation of the phenotypes of the third wave.}
\label{fig:phenotypes:wave3}

\includegraphics[width=\linewidth]{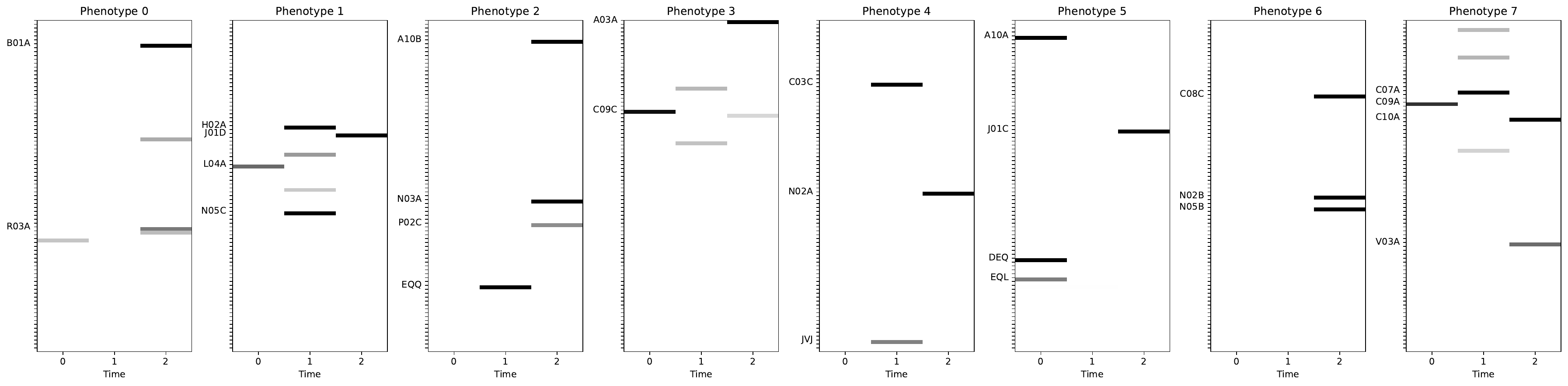}
\caption{Graphical representation of the phenotypes of the fourth wave.}
\label{fig:phenotypes:wave4}
\end{figure*}
\end{landscape}

Figure~\ref{fig:phenotypes:wave3} illustrates the phenotypes that have been extracted. These phenotypes are detailed in Table~\ref{tab:appendix:thirdwave}.

\begin{table}[p]
\scriptsize \begin{tabular}{lp{6cm}c:c:c}
\multicolumn{4}{l}{Phenotype 1}\\
\toprule
Event code & Description &  \multicolumn{3}{c}{Days}\\
\midrule\textsf{C10A} & \textit{\small Statine}& 1.00& 0.00& 0.00\\
\textsf{H02A} & \textit{\small Prednisone, antibiotic}& 1.00& 0.00& 0.00\\
\textsf{J01E} & \textit{\small Sulfonamides and trimethoprim}& 0.38& 0.00& 0.00\\
\textsf{N02A} & \textit{\small Morphines}& 0.00& 0.00& 1.00\\
\textsf{P02C} & \textit{\small Ivermectine}& 0.37& 0.00& 0.00\\
\bottomrule
\end{tabular}

\scriptsize \begin{tabular}{lp{6cm}c:c:c}
\multicolumn{4}{l}{Phenotype 2}\\
\toprule
Event code & Description &  \multicolumn{3}{c}{Days}\\
\midrule\textsf{A03A} & \textit{\small Phloroglucinol}& 0.61& 0.00& 0.00\\
\textsf{B01A} & \textit{\small Antithrombotic agents}& 1.00& 0.00& 0.00\\
\textsf{C09A} & \textit{\small Ramipril}& 0.00& 0.00& 0.66\\
\textsf{J01D} & \textit{\small Cefotaxime}& 1.00& 0.00& 0.00\\
\textsf{J01X} & \textit{\small Metronidazole}& 0.21& 0.00& 0.00\\
\bottomrule
\end{tabular}

\scriptsize \begin{tabular}{lp{6cm}c:c:c}
\multicolumn{4}{l}{Phenotype 3}\\
\toprule
Event code & Description &  \multicolumn{3}{c}{Days}\\
\midrule\textsf{A10B} & \textit{\small Metformine}& 1.00& 0.00& 0.00\\
\textsf{C07A} & \textit{\small Bisoprolol}& 0.00& 1.00& 0.00\\
\textsf{J01C} & \textit{\small Amoxicilline}& 0.00& 1.00& 0.00\\
\textsf{L04A} & \textit{\small Tocilizumab}& 0.00& 0.00& 0.51\\
\bottomrule
\end{tabular}

\scriptsize \begin{tabular}{lp{6cm}c:c:c}
\multicolumn{4}{l}{Phenotype 4}\\
\toprule
Event code & Description &  \multicolumn{3}{c}{Days}\\
\midrule\textsf{C03C} & \textit{\small Furosemide}& 0.00& 1.00& 0.00\\
\textsf{N03A} & \textit{\small Pregabaline}& 0.00& 0.88& 0.00\\
\bottomrule
\end{tabular}

\scriptsize \begin{tabular}{lp{6cm}c:c:c}
\multicolumn{4}{l}{Phenotype 5}\\
\toprule
Event code & Description &  \multicolumn{3}{c}{Days}\\
\midrule\textsf{A06A} & \textit{\small Constipation}& 0.00& 0.26& 0.00\\
\textsf{C09C} & \textit{\small Irbesartan, ARB}& 0.00& 0.40& 0.00\\
\textsf{N02B} & \textit{\small Paracetamol}& 0.00& 1.00& 0.00\\
\textsf{R03A} & \textit{\small Adrenergics, inhalants}& 0.00& 0.60& 0.00\\
\textsf{R03B} & \textit{\small Drugs for obstructive airway diseases}& 0.00& 0.27& 0.00\\
\bottomrule
\end{tabular}

\scriptsize \begin{tabular}{lp{6cm}c:c:c}
\multicolumn{4}{l}{Phenotype 6}\\
\toprule
Event code & Description &  \multicolumn{3}{c}{Days}\\
\midrule\textsf{A10A} & \textit{\small Insuline}& 0.00& 1.00& 0.00\\
\textsf{C08C} & \textit{\small Amlodipine}& 0.00& 0.00& 1.00\\
\textsf{JVJ} & \textit{\small Dialysis}& 0.00& 0.63& 0.00\\
\bottomrule
\end{tabular}

\scriptsize \begin{tabular}{lp{6cm}c:c:c}
\multicolumn{4}{l}{Phenotype 7}\\
\toprule
Event code & Description &  \multicolumn{3}{c}{Days}\\
\midrule\textsf{N05C} & \textit{\small Midazolam, sedation}& 0.00& 0.00& 1.00\\
\textsf{V03A} & \textit{\small Oxygen}& 0.00& 0.79& 0.00\\
\textsf{DEQ} & \textit{\small Electrocardiogram}& 0.00& 1.00& 0.00\\
\textsf{EQL} & \textit{\small Dopamine}& 0.00& 1.00& 0.00\\
\textsf{EQQ} & \textit{\small Cardiac monitoring}& 0.01& 1.00& 0.01\\
\textsf{YYY} & \textit{\small Rescucitation procedures}& 0.00& 0.74& 0.00\\
\bottomrule
\end{tabular}

\scriptsize \begin{tabular}{lp{6cm}c:c:c}
\multicolumn{4}{l}{Phenotype 8}\\
\toprule
Event code & Description &  \multicolumn{3}{c}{Days}\\
\midrule\textsf{N05B} & \textit{\small Hydroxizine, sedation}& 0.00& 1.00& 0.00\\
\bottomrule
\end{tabular}
\caption{Detailed description of phenotypes of the third wave.}
\label{tab:appendix:thirdwave}
\end{table}

\subsubsection{Phenotypes of the Fourth Wave}
Figure~\ref{fig:phenotypes:wave4} illustrates the phenotypes that have been extracted. These phenotypes are detailed  in Table~\ref{tab:appendix:fourthwave}.

\begin{table}[p]
\scriptsize \begin{tabular}{lp{6cm}c:c:c}
\multicolumn{4}{l}{Phenotype 1}\\
\toprule
Event code & Description &  \multicolumn{3}{c}{Days}\\
\midrule\textsf{B01A} & \textit{\small Antithrombotic agents}& 0.00& 0.00& 1.00\\
\textsf{J01E} & \textit{\small Sulfonamides and trimethoprim}& 0.00& 0.00& 0.33\\
\textsf{R03A} & \textit{\small Adrenergics, inhalants}& 0.00& 0.00& 0.53\\
\textsf{R03B} & \textit{\small Drugs for obstructive airway diseases}& 0.00& 0.00& 0.26\\
\textsf{R06A} & \textit{\small Antihistamines}& 0.23& 0.00& 0.00\\
\bottomrule
\end{tabular}

\scriptsize \begin{tabular}{lp{6cm}c:c:c}
\multicolumn{4}{l}{Phenotype 2}\\
\toprule
Event code & Description &  \multicolumn{3}{c}{Days}\\
\midrule\textsf{H02A} & \textit{\small Prednisone, antibiotic}& 0.00& 1.00& 0.00\\
\textsf{J01D} & \textit{\small Cefotaxime}& 0.00& 0.00& 1.00\\
\textsf{J01X} & \textit{\small Metronidazole}& 0.00& 0.40& 0.00\\
\textsf{L04A} & \textit{\small Tocilizumab}& 0.59& 0.00& 0.00\\
\textsf{N01B} & \textit{\small Lidocaine}& 0.00& 0.21& 0.00\\
\textsf{N05C} & \textit{\small Midazolam, sedation}& 0.00& 1.00& 0.00\\
\bottomrule
\end{tabular}

\scriptsize \begin{tabular}{lp{6cm}c:c:c}
\multicolumn{4}{l}{Phenotype 3}\\
\toprule
Event code & Description &  \multicolumn{3}{c}{Days}\\
\midrule\textsf{A10B} & \textit{\small Metformine}& 0.00& 0.00& 1.00\\
\textsf{N03A} & \textit{\small Pregabaline}& 0.00& 0.00& 1.00\\
\textsf{P02C} & \textit{\small Ivermectine}& 0.00& 0.00& 0.45\\
\textsf{EQQ} & \textit{\small Cardiac monitoring}& 0.00& 1.00& 0.00\\
\bottomrule
\end{tabular}

\scriptsize \begin{tabular}{lp{6cm}c:c:c}
\multicolumn{4}{l}{Phenotype 4}\\
\toprule
Event code & Description &  \multicolumn{3}{c}{Days}\\
\midrule\textsf{A03A} & \textit{\small Phloroglucinol}& 0.00& 0.00& 1.00\\
\textsf{C03D} & \textit{\small Spironolactone, cardiac failure}& 0.00& 0.28& 0.00\\
\textsf{C09C} & \textit{\small Irbesartan, ARB}& 0.94& 0.00& 0.00\\
\textsf{C09D} & \textit{\small Anti-hypertensor ARB}& 0.00& 0.00& 0.16\\
\textsf{J01F} & \textit{\small Azithromicine}& 0.00& 0.24& 0.00\\
\bottomrule
\end{tabular}

\scriptsize \begin{tabular}{lp{6cm}c:c:c}
\multicolumn{4}{l}{Phenotype 5}\\
\toprule
Event code & Description &  \multicolumn{3}{c}{Days}\\
\midrule\textsf{C03C} & \textit{\small Furosemide}& 0.00& 1.00& 0.00\\
\textsf{N02A} & \textit{\small Morphines}& 0.00& 0.00& 1.00\\
\textsf{JVJ} & \textit{\small Dialysis}& 0.00& 0.49& 0.00\\
\bottomrule
\end{tabular}

\scriptsize \begin{tabular}{lp{6cm}c:c:c}
\multicolumn{4}{l}{Phenotype 6}\\
\toprule
Event code & Description &  \multicolumn{3}{c}{Days}\\
\midrule\textsf{A10A} & \textit{\small Insuline}& 1.00& 0.00& 0.00\\
\textsf{J01C} & \textit{\small Amoxicilline}& 0.00& 0.00& 1.00\\
\textsf{DEQ} & \textit{\small Electrocardiogram}& 1.00& 0.00& 0.00\\
\textsf{EQL} & \textit{\small Dopamine}& 0.50& 0.00& 0.00\\
\bottomrule
\end{tabular}

\scriptsize \begin{tabular}{lp{6cm}c:c:c}
\multicolumn{4}{l}{Phenotype 7}\\
\toprule
Event code & Description &  \multicolumn{3}{c}{Days}\\
\midrule\textsf{C08C} & \textit{\small Amlodipine}& 0.00& 0.00& 1.00\\
\textsf{N02B} & \textit{\small Paracetamol}& 0.00& 0.00& 1.00\\
\textsf{N05B} & \textit{\small Hydroxizine, sedation}& 0.00& 0.00& 1.00\\
\bottomrule
\end{tabular}

\scriptsize \begin{tabular}{lp{6cm}c:c:c}
\multicolumn{4}{l}{Phenotype 8}\\
\toprule
Event code & Description &  \multicolumn{3}{c}{Days}\\
\midrule\textsf{A06A} & \textit{\small Constipation}& 0.00& 0.27& 0.00\\
\textsf{C01B} & \textit{\small Amiodarone, cardiac failure}& 0.00& 0.29& 0.00\\
\textsf{C07A} & \textit{\small Bisoprolol}& 0.00& 1.00& 0.00\\
\textsf{C09A} & \textit{\small Ramipril}& 0.81& 0.00& 0.00\\
\textsf{C10A} & \textit{\small Statine}& 0.00& 0.00& 1.00\\
\textsf{J01M} & \textit{\small Ciprofloxacine}& 0.00& 0.18& 0.00\\
\textsf{V03A} & \textit{\small Oxygen}& 0.00& 0.00& 0.58\\
\bottomrule
\end{tabular}
\caption{Detailed description of phenotypes of the fourth wave.}
\label{tab:appendix:fourthwave}
\end{table}

\end{appendices}

\end{document}